\documentclass{article}
\PassOptionsToPackage{numbers,compress}{natbib}

\usepackage[preprint]{neurips_2019}

\usepackage[utf8]{inputenc} 
\usepackage[T1]{fontenc}    
\usepackage{hyperref}       
\usepackage{url}            
\usepackage{booktabs}       
\usepackage{amsfonts}       
\usepackage{nicefrac}       
\usepackage{microtype}      
\usepackage{amsmath,amsthm,amssymb}
\usepackage{bm}
\usepackage{color}
\usepackage{dsfont}

\usepackage{xspace}
\usepackage{url}
\usepackage{makecell}

\usepackage{graphicx}
\usepackage{subcaption}
\usepackage{booktabs} 
\usepackage{algorithm}
\usepackage{algorithmic}
\usepackage[all]{nowidow}
\usepackage{enumitem}

\setlist[enumerate]{itemsep=0.5pt, wide=\parindent}

\theoremstyle{definition}

\def\0{{\bm 0}}

\def\b{{\bm b}}

\def\s{{\bm s}}

\def\v{{\bm v}}

\def\B{{\bm B}}
\def\C{{\bm C}}

\def\I{{\bm I}}

\def\L{{\bm L}}

\def\V{{\bm V}}

\def\Z{{\bm Z}}

\def\Lambda{\boldsymbol{\lambda}}

\def\Acal{\mathcal{A}}

\def\Pcal{\mathcal{P}}

\def\Tcal{\mathcal{T}}

\def\Ycal{\mathcal{Y}}

\def\Rbb{\mathbb{R}}

\def\wo{\backslash}

\title{Deep Determinantal Point Processes}

\author{%
  Mike Gartrell \\
  Criteo AI Lab \\
  \texttt{m.gartrell@criteo.com} \\
  \And
  Elvis Dohmatob \\
  Criteo AI Lab \\
  \texttt{e.dohmatob@criteo.com} \\
  \And
  Jon Alberdi \\
  Criteo \\
  \texttt{j.alberdi@criteo.com} \\
}

\begin{document}

\maketitle

\begin{abstract}
Determinantal point processes (DPPs) have attracted significant attention as an
elegant model that is able to capture the balance between quality and diversity
within sets.  DPPs are parameterized by a positive semi-definite kernel matrix.
While DPPs have substantial expressive power, they are fundamentally limited by
the parameterization of the kernel matrix and their inability to capture
nonlinear interactions between items within sets.  We present the \emph{deep
DPP} model as way to address these limitations, by using a deep feed-forward
neural network to learn the kernel matrix.  In addition to allowing us to
capture nonlinear item interactions, the deep DPP also allows easy incorporation
of item metadata into DPP learning.  Since the learning target is the DPP kernel
matrix, the deep DPP allows us to use existing DPP algorithms for efficient
learning, sampling, and prediction.  Through an evaluation on several real-world
datasets, we show experimentally that the deep DPP can provide a considerable
improvement in the predictive performance of DPPs, while also outperforming
strong baseline models in many cases. 
\end{abstract}
\vspace{-0.3cm}
\section{Introduction}
\vspace{-0.3cm}
Modeling the relationship between items within observed subsets, drawn from a
large collection, is an important challenge that is fundamental to many machine
learning applications, including recommender systems~\cite{gillenwater-thesis},
document summarization~\cite{kulesza2011learning, lin12}, and information
retrieval~\cite{kulesza11}.  For these applications, we are primarily concerned
with selecting a good subset of diverse, high-quality items.  Balancing quality
and diversity in this setting is challenging, since the number of possible
subsets that could be drawn from a collection grows exponentially as the
collection size increases.

Determinantal point processes (DPPs) offer an elegant and attractive model for
such tasks, since they provide a tractable model that jointly considers set
diversity and item quality.  A DPP models a distribution over subsets of a
ground set $\Ycal$ that is parametrized by a positive semi-definite matrix $\L \in
\Rbb^{|\Ycal| \times |\Ycal|}$, such that for any $A \subseteq \mathcal Y$,
\begin{equation}
  \Pr(A) \propto \det(\L_A),
  \label{eq:dpp}
\end{equation}
where $\L_A=[\L_{ij}]_{i,j\in A}$ is the submatrix of $\L$ indexed by $A$.
Informally, $\det(\L_A)$ represents the volume associated with subset $A$, the
diagonal entry $L_{ii}$ represents the importance of item $i$, while entry
$L_{ij} = L_{ji}$ encodes the similarity between items $i$ and $j$.  DPPs have
been studied in the context of a number of applications~\cite{affandi14,chao15,
krause08,mariet16,zhang17} in addition to those mentioned above.  There has
also been significant work regarding the theoretical properties of
DPPs~\cite{kuleszaBook,borodin2009,affandi14,kuleszaThesis,gillenwater-thesis,decreuse,lavancier15}.

Learning a DPP from observed data in the form of example subsets is a
challenging task that is conjectured to be NP-hard~\cite{kuleszaBook}.  Some
work has involved learning a nonparametric full-rank $\L$
matrix~\cite{gillenwater-thesis, mariet15} that does not constrain $\L$ to take
a particular parametric form, while other work has involved learning a low-rank
factorization of this nonparametric $\L$ matrix~\cite{gartrell17, osogami18}. A
low-rank factorization of $\L$ enables substantial improvements in runtime
performance compared to a full-rank DPP model during training and when computing
predictions, on the order of 10-20x or more, with predictive performance that is
equivalent to or better than a full-rank model.

While the low-rank DPP model scales well, it has a fundamental limitation
regarding model capacity and expressive power due to the nature of the low-rank
factorization of $\L$.  A rank-$K$ factorization of $\L$ has an implicit
constraint on the space of possible subsets, since it places zero probability
mass on subsets with more than $K$ items. When trained on a dataset containing
subsets with at most $K$ items, we observe from the results in~\cite{gartrell17}
that this constraint is reasonable and that the rank-$K$ DPP provides predictive
performance that is approximately equivalent to that of the full-rank DPP.
Therefore, in this scenario the rank-$K$ DPP can be seen as a good approximation
of the full-rank DPP. However, we empirically observe that the rank-$K$ DPP
generally does not provide improved predictive performance for values of $K$
greater than the size of the largest subset in the data. Thus, for a dataset
containing subsets no larger than size $K$, from the standpoint of predictive
performance, there is generally no utility in increasing the number of low-rank
DPP embedding dimensions beyond $K$, which establishes an upper bound on the
capacity of the model.  Furthermore, since the determinant is a multilinear
function of the columns or rows of a matrix, a DPP is unable to capture
nonlinear interactions between items within observed subsets.

The constraints of the standard DPP model motivate us to seek modeling options
that enable us to increase the expressive power of the model and improve
predictive performance, while still allowing us to leverage the efficient
learning, sampling, and prediction algorithms available for DPPs.  We present
the \emph{deep DPP} as a model that fulfills these requirements. The deep DPP
uses a deep feed-forward neural network to learn the low-rank DPP embedding
matrix, allowing us to move beyond the constraints of the standard multilinear
DPP model by supporting nonlinearities in the embedding space through the use of
multiple hidden layers in the deep network.  The deep DPP also allows us to
incorporate item-level metadata into the model, such as item names,
descriptions, etc. Since the learning target of the deep DPP model is the
low-rank DPP embedding matrix, we can use existing algorithms for efficient
learning, sampling, and prediction for DPPs.  Thus, the deep DPP provides us
with an elegant deep generative model for sets.

\paragraph*{Contributions}The main contributions of this work are the following:
\begin{itemize}
    \item We extend the standard low-rank DPP model by using a deep feed-forward
    neural network for learning the DPP kernel matrix, which is composed of item
    embeddings.  This approach allows us to arbitrarily increase the expressive
    power of the deep DPP model by simply adding hidden layers to the deep
    network.
    \item The deep DPP supports arbitrary item-level metadata.  The deep network
    in our model allows us to easily incorporate such metadata, and
    automatically learns parameters that explain how this metadata interacts
    with the latent item embeddings in our model.  In recommendation settings,
    leveraging metadata has been shown to improve predictive
    quality~\cite{kula2015metadata, vasile2016meta}, particularly for cold-start
    scenarios with sparse user/item interactions.
    \item We conduct an extensive experimental evaluation on several real-world
    datasets.  This analysis highlights scenarios in which the deep DPP can
    provide significantly better predictive quality compared to the standard
    low-rank DPP.  Specifically, we see that the deep DPP is able to extract
    complex nonlinear item interactions, particularly for large, complex
    datasets.
\end{itemize}
\section{Related Work}
\label{sec:related-work}

A number of approaches for DPP kernel learning have been studied.
~\cite{gillenwater14} presents DPP kernel learning via expectation maximization,
while ~\cite{mariet15} present a fixed-point method.  Methods to substantially
speed up DPP kernel learning, as compared to learning a full-rank kernel, have
leveraged Kronecker~\cite{mariet16b} and
low-rank~\cite{dupuy16,gartrell2016bayesian, gartrell17,osogami18} structures.
~\cite{mariet2019negdpp} presents methods for incorporating inferred negative
samples into the DPP learning task, based on contrastive estimation.  Learning
guarantees using DPP graph properties are studied in~\cite{urschel17}.

There has been some prior work regarding the use of a deep neural network to
learn DPP model parameters.  In~\cite{xie2017deep}, the authors describe one
approach that involves a deep network, where the model is parameterized in terms
of one data instance (an image) for each observed subset of labels.  In
contrast, our deep DPP model is more general and allows for each item within the
ground set $\Ycal$ to have its own item-level metadata features (item price,
description, image, etc.).  ~\cite{wilhelm2018practical} present an approach
for generating YouTube video recommendations, where two different deep networks
are used to learn a decomposition of the DPP $\L$ matrix based on item quality
scores and item embedding vectors.  Our deep DPP approach differs, in that we
implement end-to-end learning of a fully non-parametric $\L$ that does not
impose any particular parametric form.  ~\cite{mariet2019dppnet} presents 
a deep generative model that efficiently generates samples which approximate 
samples drawn from a DPP.  However, this approach does not involve using a deep
network to learn DPP model parameters.

Matrix factorization models are commonly used to model user-item interactions in
recommender systems.  Although DPP and matrix factorization models are
fundamentally different, they are both limited to capturing linear interactions
between items or users and items, respectively.  An approach to extending the
conventional matrix factorization model by using a deep network to learn
representations of user and items within a nonlinear embedding space is
described in~\cite{xue2017deep}.  This deep matrix factorization model bears
some conceptual similarity with our deep DPP model, in that both approaches
involve extending linear models to accommodate nonlinear interactions by using
deep networks.

~\cite{covington2016deep} describes a YouTube recommendation system that uses
deep networks for candidate generation and ranking.  Features such as user age
and gender are concatenated with embeddings for users' watch and search
histories to form the first layer of the deep network used for the candidate
generation model, and each hidden layer is fully connected.  This network
architecture bears some similarity to the architecture we use to support both
item embeddings and metadata in our deep DPP model, as described in 
Section~\ref{sec:model}.

\vspace{-0.3cm}
\section{Model}
\label{sec:model}
\vspace{-0.3cm}

\begin{figure*}[t!]
    \centering
    \includegraphics[width=0.80\textwidth]{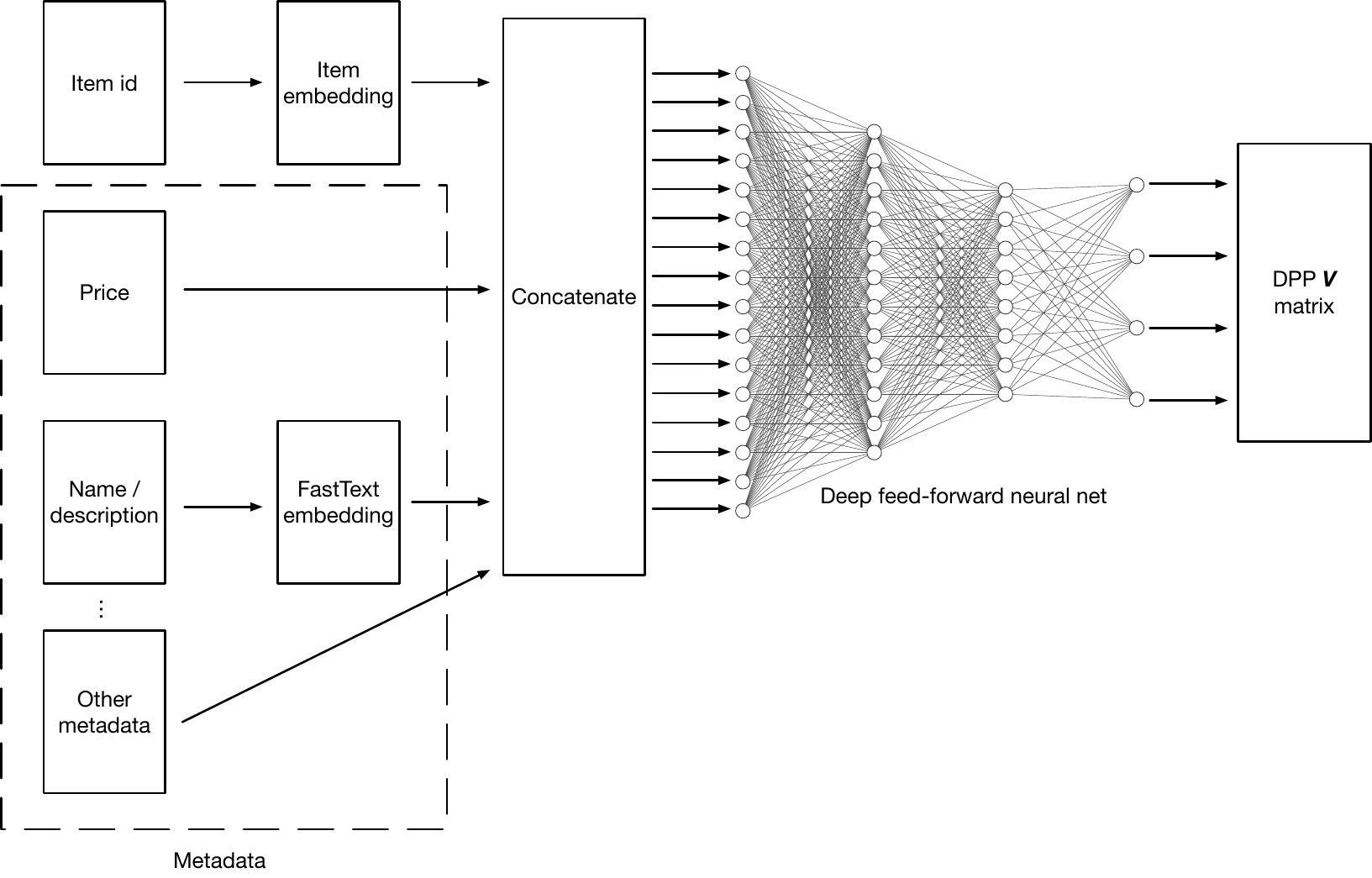}
    \caption{Deep DPP architecture.  Observed subsets composed of item ids, as
    well as optional item-level metadata, are provided as inputs to the deep
    neural network during learning.  The output is the learned $|\Ycal| \times
    K$ DPP parameter matrix $\V$, where each row of this matrix is an item
    embedding vector.}
    \vspace{-0.5cm}
\label{fig:architecture}
\end{figure*}

We begin this section with some background on DPPs and low-rank DPPs, followed
by a discussion of the architecture of our deep DPP model.

Since the normalization constant for Eq.~\ref{eq:dpp} follows from
the observation that $\sum_{A\subseteq \Ycal}\det(\L_A)=\det(\L+\I)$,
we have 
\begin{equation}
    \Pcal(A) = \frac{\det(\L_A)}{\det(\L+\I)}
    \label{eq:dpp-normalized}
\end{equation}
where a discrete DPP is a probability measure $\Pcal$ on $2^\Ycal$ (the power set
or set of all subsets of $\Ycal$).  Therefore, the probability $\Pcal(A)$ for any 
$A \subseteq \mathcal Y$ is given by Eq.~\ref{eq:dpp-normalized}.

We use a low-rank factorization of the $|\Ycal| \times |\Ycal|$ $\L$ matrix, $\L
= \V \V^T$, where $\V \in \Rbb^{|\Ycal| \times K}$, and $K \leq |\Ycal|$ is the
rank of the kernel. $K$ is fixed \emph{a priori}, and is often set to the size
of the largest observed subset in the data.

Given a collection of $N$ observed subsets $\Acal = \{ A_1, ..., A_N \}$
composed of items from $\Ycal$, our learning task is to fit a DPP kernel $\L$
based on this data.  Our training data is these observed subsets $\Acal$, 
and our task is to maximize the likelihood for samples drawn from the same 
distribution as $\Acal$.  The log-likelihood for seeing $\Acal$ is
\begin{align}
    f(\V) & = \log \Pcal(\Acal | \V) = \sum_{n=1}^N \log \Pcal(A_n | \V) \nonumber = \sum_{n=1}^N \log \det (\L_{[n]}) - N \log \det(\L + \I)
    \label{eq:log-likelihood}
\end{align}
where $[n]$ indexes the observed subsets in $\Acal$.

As described in~\cite{gartrell17}, we augment $f(\V$) with a regularization
term:
\begin{equation}
    f(\V) = \sum_{n=1}^N \log \det (\L_{[n]}) - N \log \det(\L + \I) - \alpha \sum_{i=1}^{|\Ycal|} \frac{1}{\lambda_i} \|\v_i\|_2^2
    \label{eq:reg-log-likelihood}
\end{equation}
where $\lambda_i$ counts the number of occurrences of item $i$ in the training
set, $\v_i$ is the corresponding row vector of $\V$, and $\alpha > 0$ is a
tunable hyperparameter. This regularization term reduces the magnitude of
$\|\v_i\|_2$, which can be interpreted as the popularity of item $i$, according
to its empirical popularity $\lambda_i$.

Figure~\ref{fig:architecture} shows the architecture of the deep DPP model.  As
shown in this figure, a deep network is used to learn $\V$.  Furthermore, this
architecture allows us to seamlessly incorporate item metadata, such as price
and item name and description, into the model.  As shown in
Figure~\ref{fig:architecture}, we use FastText
embeddings~\cite{bojanowski2017enriching} to support text-based metadata, such
as item names and descriptions. We use self-normalizing SELU activation
functions~\cite{klambauer2017self} for our deep network, since we empirically
found that this activation function provides stable convergence behavior during
training.  The network is structured according to a common tower-like pattern,
where the first layer is widest, and each of the following hidden layers reduces
the number of hidden units, until we reach the target embedding size $K$.  For
example, a network with three hidden layers for a model for a catalog of 50,000
items and a target embedding size of $K=100$, not considering metadata, would
use the following architecture: $50,000 \rightarrow 400 \rightarrow \text{SELU}
\rightarrow 300 \rightarrow \text{SELU} \rightarrow 200 \rightarrow \text{SELU}
\rightarrow 100$.

We use the Adam stochastic optimization algorithm~\cite{kingma2015adam} to train
our model, in conjunction with Hogwild~\cite{recht2011hogwild} for asynchronous
parallel updates during training.  Algorithm~\ref{alg:learning-algorithm} shows
the learning algorithm for our model.  All code is implemented in
PyTorch~\footnote{\url{https://pytorch.org}}, and will be made publicly
available at a later date.

\begin{algorithm}[t]
    \caption{Learning the deep DPP $\V$ matrix}
    \label{alg:deep-dpp-learning}
    \begin{algorithmic}
        \STATE {\bfseries Input:} Samples of training subsets $\Acal$, initial parameter matrix $\V_0$, maxIter, deepNetArch.
        \STATE \textsc{BuildDeepNet}(deepNetArch, $\V$)
        \STATE $k \leftarrow 1$
        \WHILE {$k<$ maxIter {\bfseries and} not converged}
        \STATE \textbf{Compute embeddings} matrix $\V_k$ via forward pass
        \STATE \textbf{Sample mini-batch} of baskets and evaluate loss $f(\V_k, \Acal)$
        \STATE $\V_{k+1} \leftarrow \textbf{Backrop}$ on loss
        \ENDWHILE
        \item[] \textbf{return} $\V_k$
    \end{algorithmic}
    \label{alg:learning-algorithm}
    \vspace{-0.1cm}
\end{algorithm}

\section{Experiments}
\label{sec:experiments}

\begin{figure*}[t]
    \centering
    \begin{subfigure}{.40\textwidth}
        \centering
        \includegraphics[width=\textwidth]{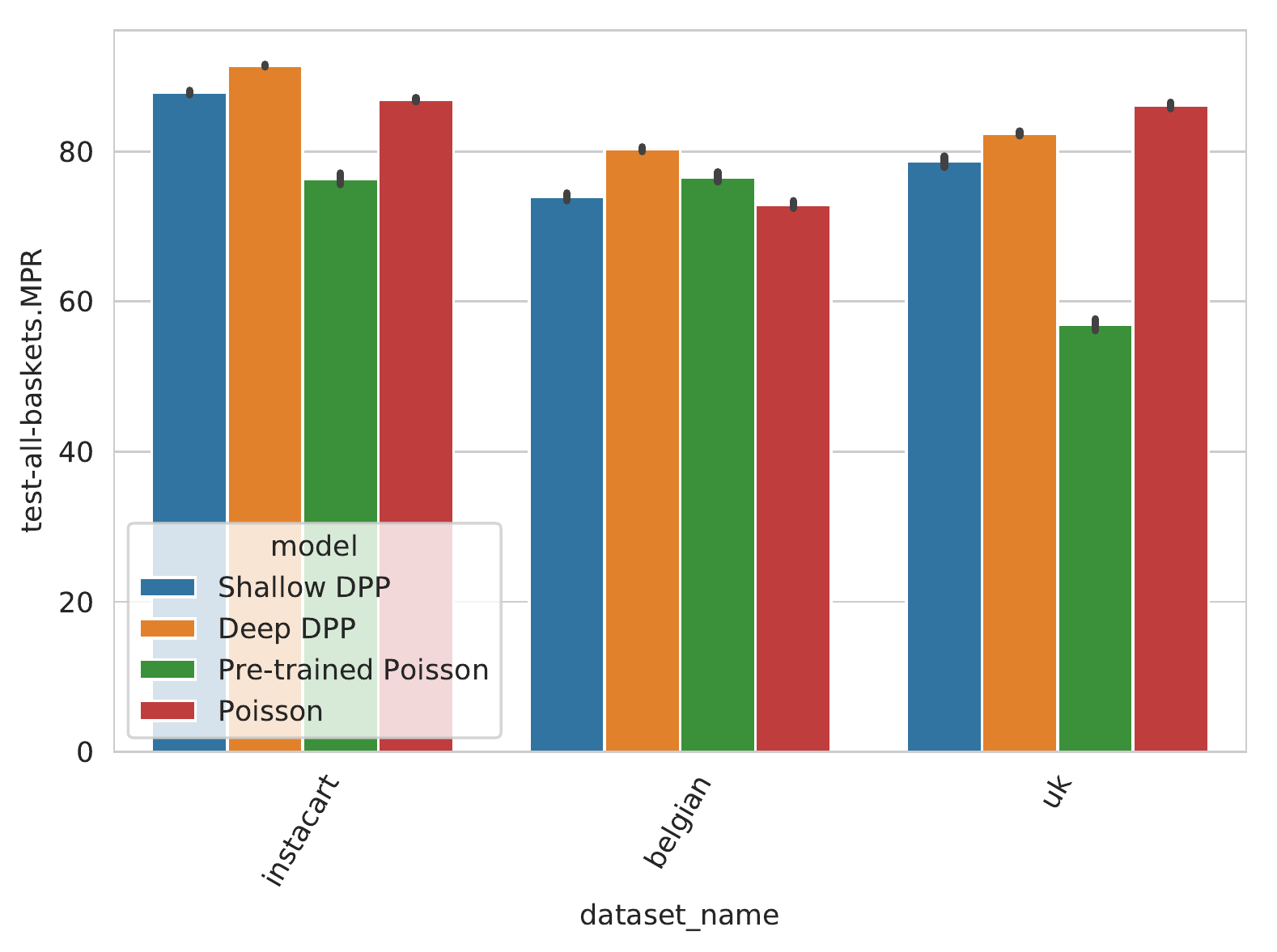}
    \end{subfigure}
    \begin{subfigure}{.40\textwidth}
        \centering
        \includegraphics[width=\textwidth]{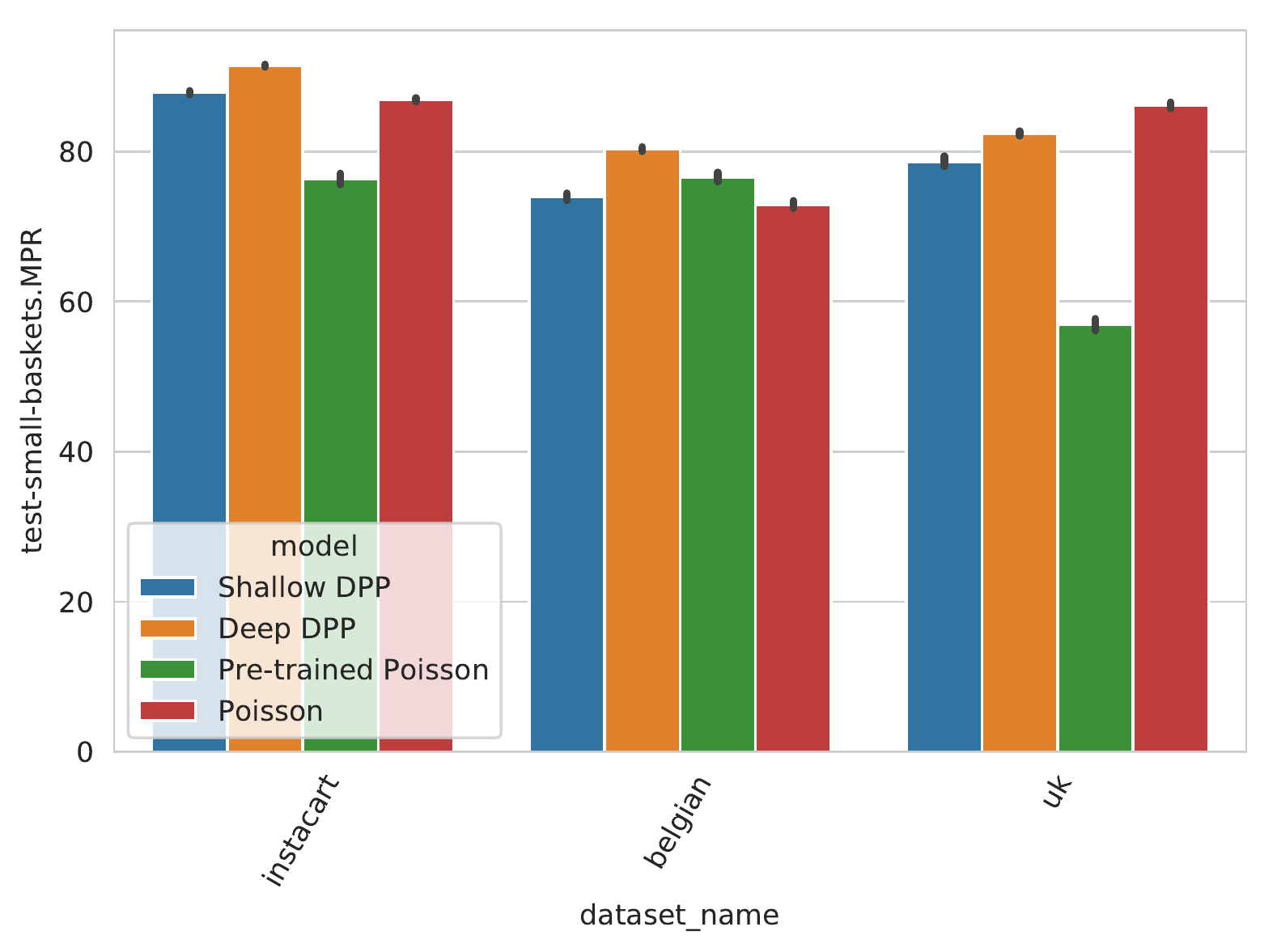}
    \end{subfigure}
    \begin{subfigure}{.40\textwidth}
        \centering
        \includegraphics[width=\textwidth]{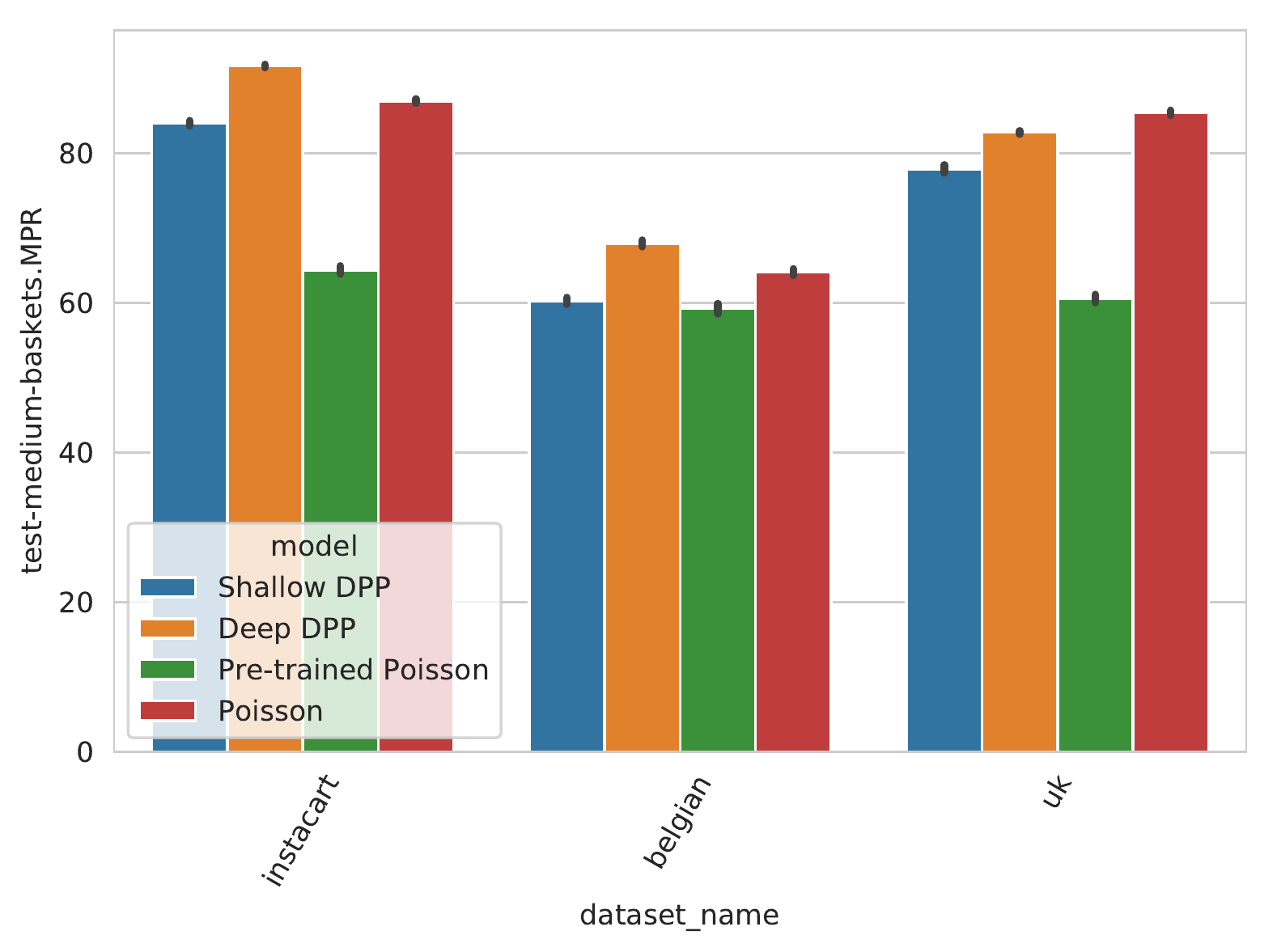}
    \end{subfigure}
    \begin{subfigure}{.40\textwidth}
        \centering
        \includegraphics[width=\textwidth]{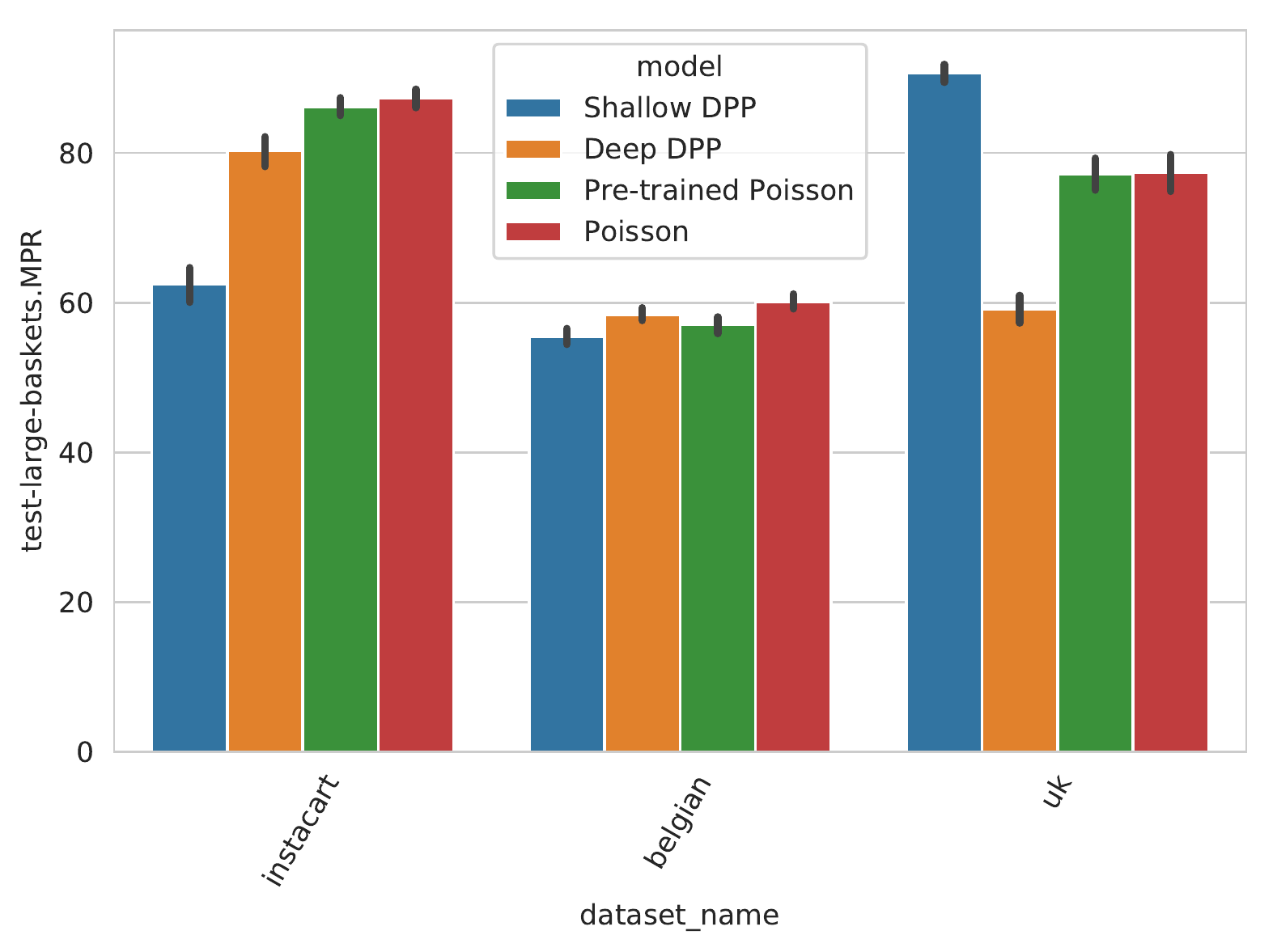}
    \end{subfigure}
    \vspace{-0.2cm}
    \caption{MPR results for the Instacart, Belgian, and UK datasets.  Metadata 
    is not used for any of these models.}
    \label{fig:MPR-results}
    \vspace{-0.4cm}
\end{figure*}

\begin{figure*}[t]
    \centering
    \begin{subfigure}{.40\textwidth}
        \centering
        \includegraphics[width=\textwidth]{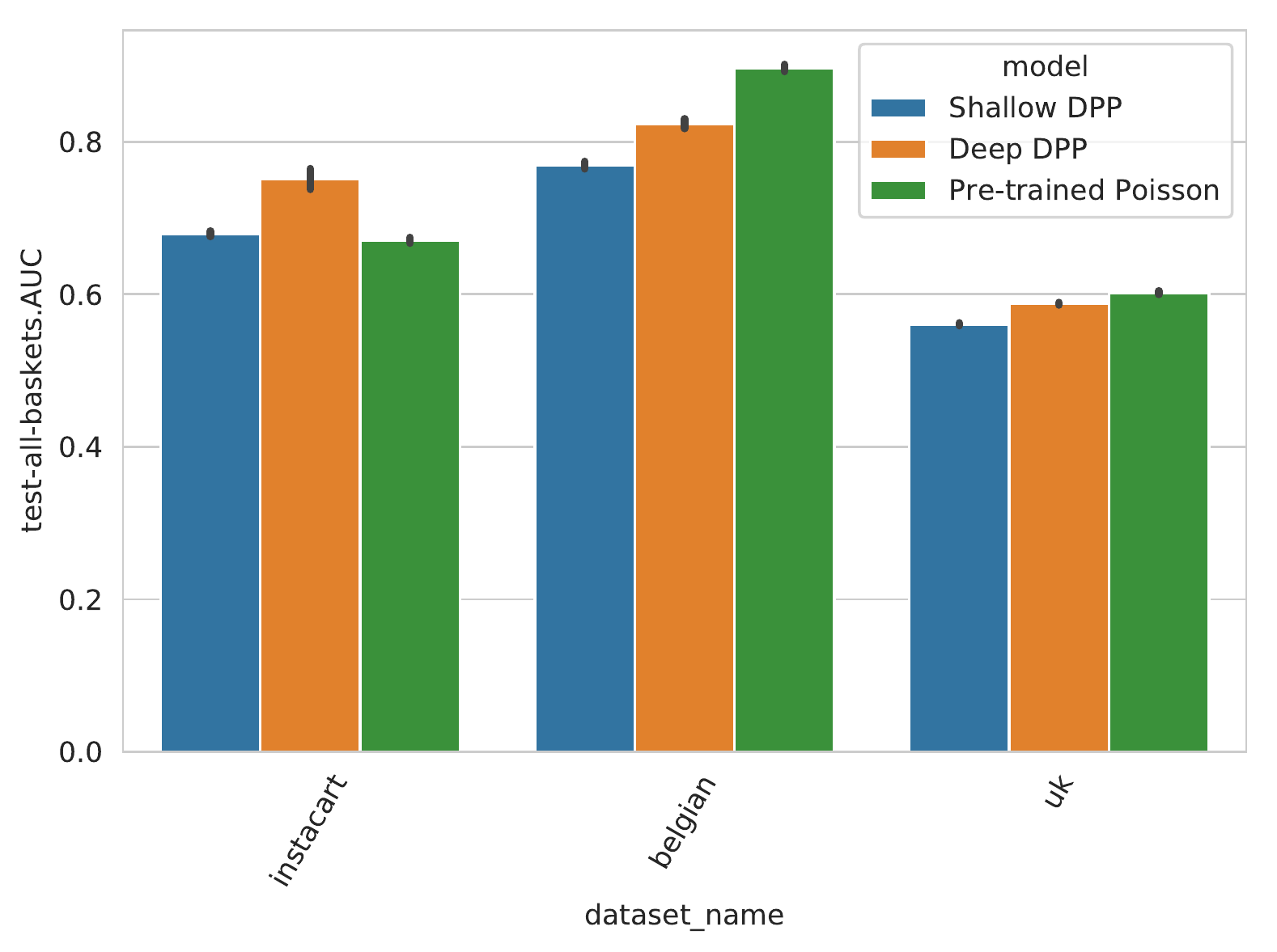}
    \end{subfigure}
    \begin{subfigure}{.40\textwidth}
        \centering
        \includegraphics[width=\textwidth]{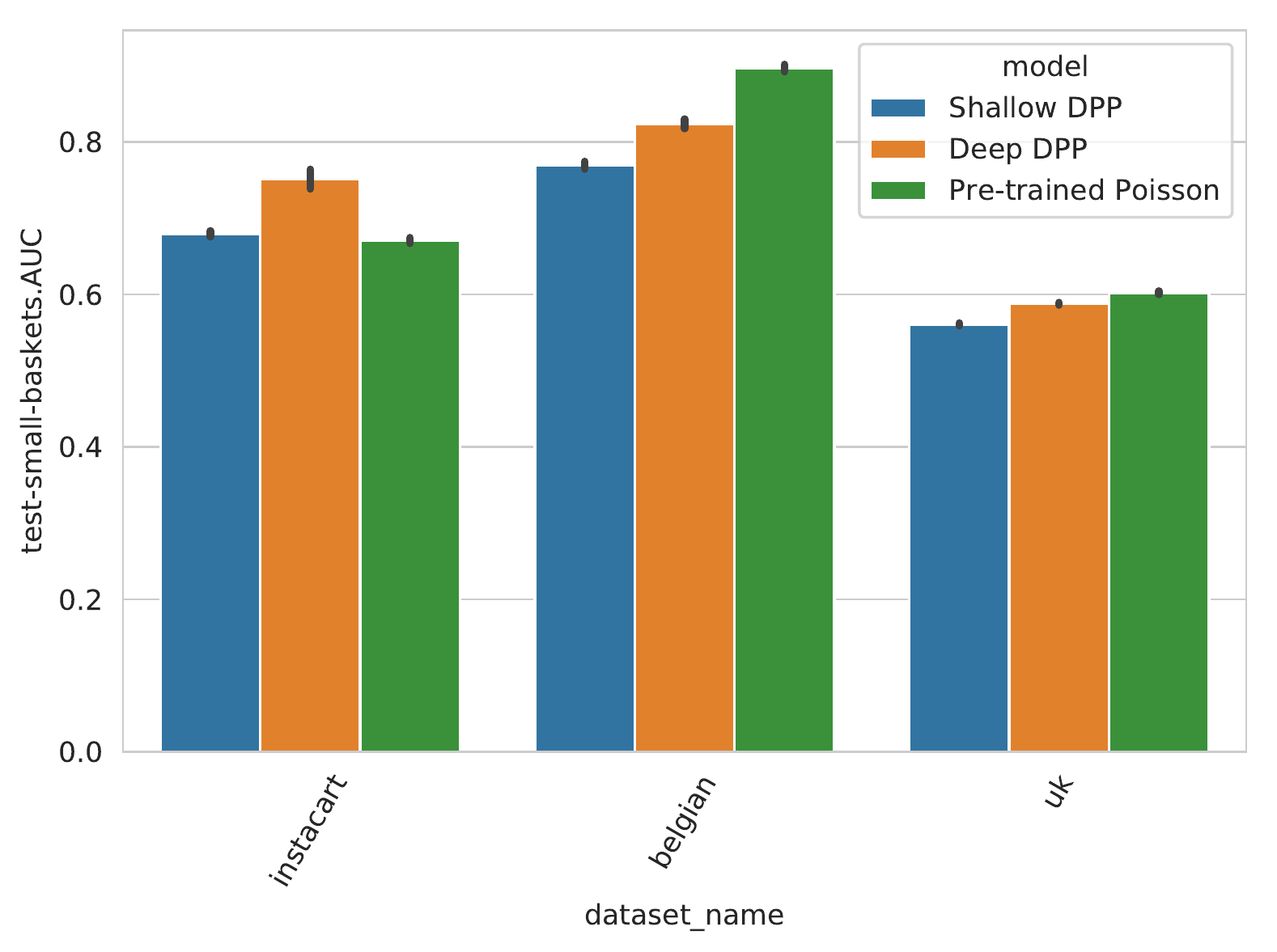}
    \end{subfigure}
    \begin{subfigure}{.40\textwidth}
        \centering
        \includegraphics[width=\textwidth]{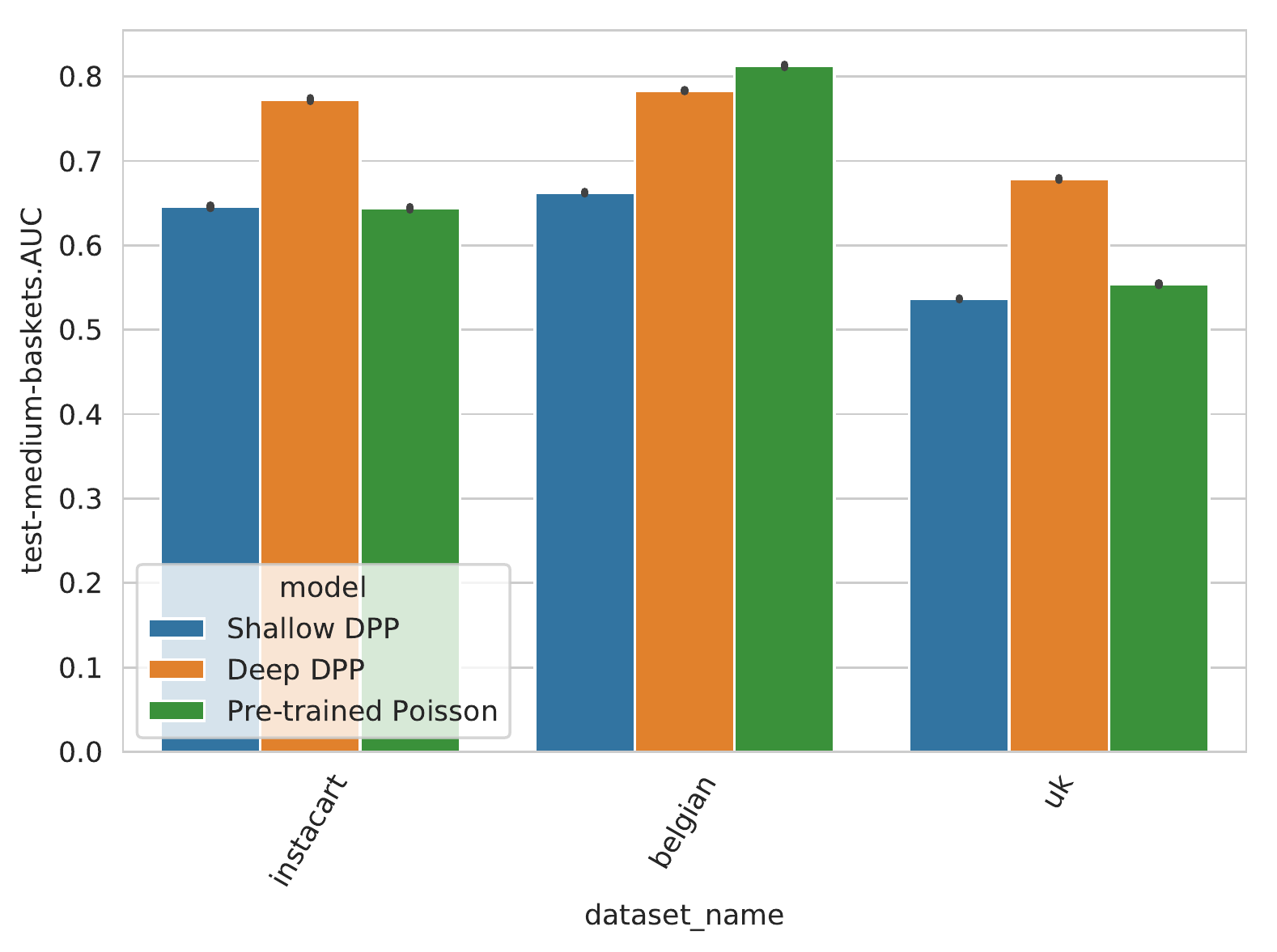}
    \end{subfigure}
    \begin{subfigure}{.40\textwidth}
        \centering
        \includegraphics[width=\textwidth]{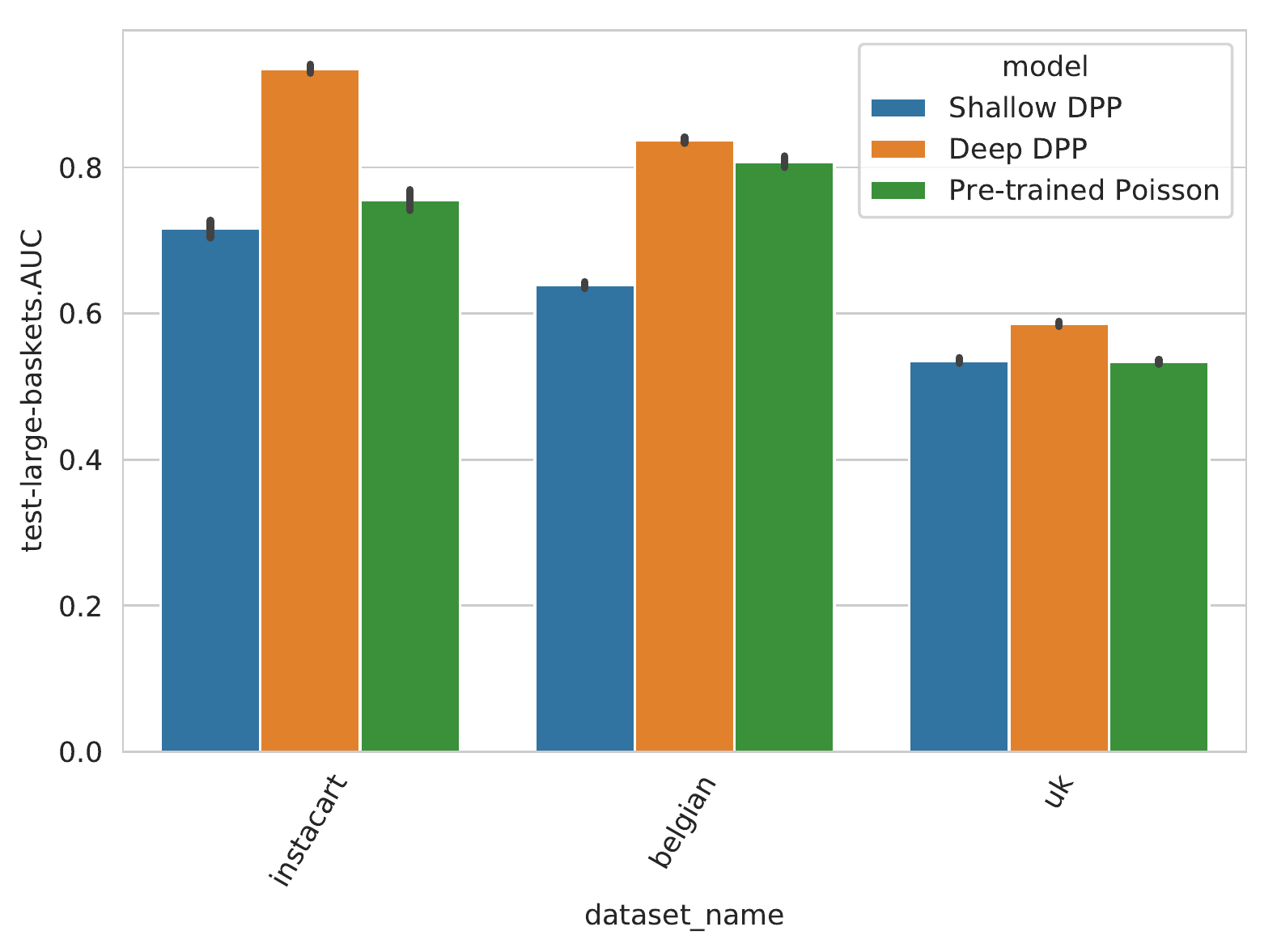}
    \end{subfigure}
    \vspace{-0.2cm}
    \caption{AUC results results for the Instacart, Belgian, and UK datasets.  Metadata 
    is not used for any of these models.}
    \label{fig:AUC-results}
    \vspace{-0.4cm}
\end{figure*}

We run extensive experiments on several real-world datasets composed of shopping
baskets.  The primary prediction task we evaluate is next-item prediction, which
involves identifying the best item to add to a subset of chosen objects (e.g.,
basket completion); see Appendix~\ref{section:computing-predictions} for details
on how we efficiently compute such predictions for the low-rank DPP and deep
DPP models.  We compare our deep DPP model to the following competing approaches:

\begin{enumerate}
    \item \textbf{Low-rank DPP:} The standard low-rank DPP
    model~\cite{gartrell17}. 
    \item \textbf{Poisson Factorization (PF):} Poisson factorization (PF) is a
    prominent variant of matrix factorization, designed specifically for
    implicit ratings (e.g., clicks or purchase events)~\cite{gopalan2015}.
    Since PF is parameterized in terms of users and items, rather than subsets,
    we train PF by treating each observed basket as a "user".  When computing
    next-item predictions for baskets, inspired by the approach for set
    expansion described in~\cite{zaheer2017deep}, we average the embeddings for
    each item within the observed basket; let us denote the resulting embedding
    vector for the basket as $\s_A$.  Then, we compute the unnormalized
    probability of each possible next item for the basket, which is proportional
    to the inner product between $\s_A$ and the embedding for the next item.  We
    use a publicly available implementation of
    PF~\footnote{\url{https://github.com/david-cortes/hpfrec}}, with default
    hyperparameter settings, and 30 embedding dimensions.
    \item \textbf{Low-rank DPP constructed from pre-trained PF embeddings:} We
    examine the value of end-to-end learning of the deep DPP kernel, compared to
    a low-rank DPP kernel directly constructed from embeddings learned using a
    different model. Here, we simply set the low-rank DPP $\V$ matrix to the
    matrix of item embeddings obtained from the PF model, rather than learning
    the $\V$ matrix using the DPP log-likelihood
    (Eq.~\ref{eq:reg-log-likelihood}).  We refer to this model as the
    "pre-trained Poisson" model in Figures~\ref{fig:MPR-results}
    and~\ref{fig:AUC-results}.
\end{enumerate} 

\vspace{-0.2cm}
\subsection{Datasets}
\label{subsec:experimental-datasets}

We perform next-item prediction and AUC-based classification experiments on
several real-world datasets composed of purchased shopping baskets:
\begin{enumerate}
    \item \textbf{UK Retail:} This is a public dataset~\cite{lsbupr1492}
    that contains 25,898 baskets drawn from a catalog of 4,070 items, and provides
    price and description metadata for each item, which we use in our
    experiments.  This dataset contains transactions from a non-store online
    retail company that primarily sells unique all-occasion gifts, and many
    customers are wholesalers.  We omit all baskets with more than 100 items,
    which allows us to use a low-rank factorization of the DPP ($K = 100$) that
    scales well in training and prediction time, while also keeping memory
    consumption for model parameters to a manageable level.
    \item \textbf{Belgian Retail Supermarket:} This public
    dataset\footnote{\url{http://fimi.ua.ac.be/data/retail.pdf}} includes 88,163
    baskets, with a catalog consisting of 16,470 unique supermarket items.  This
    dataset was collected in a Belgian retail supermarket over three
    non-consecutive time periods~\cite{brijs99,brijs03}.  We set $K = 100$ for
    all DPP models trained on this dataset, to accommodate the largest basket
    found in this datset.
    \item \textbf{Instacart:}  This public dataset~\footnote{\url{https://www.instacart.com/datasets/grocery-shopping-2017}}
    is composed of 3.2 million baskets purchased by more than 200,000 users of
    the Instacart service, drawn from a catalog of 49,677 products.  We use
    metadata provided by this dataset that includes supermarket department ID,
    aisle ID, and product name for each product in our experiments.  As with the
    UK retail dataset, we omit all baskets with more than 100 items.  In addition
    to running experiments on the full dataset, we run experiments on a random
    sample of 10,000 baskets from the full dataset, with a catalog of 16,258
    products for this random sample; we denote this smaller dataset as
    \emph{Instacart-10k} in the presentation of our experimental results in
    this paper, while the full dataset is denoted as \emph{Instacart}.
\end{enumerate}

\begin{figure*}[t]
    \centering
    \begin{subfigure}{.24\textwidth}
        \centering
        \includegraphics[width=\textwidth]{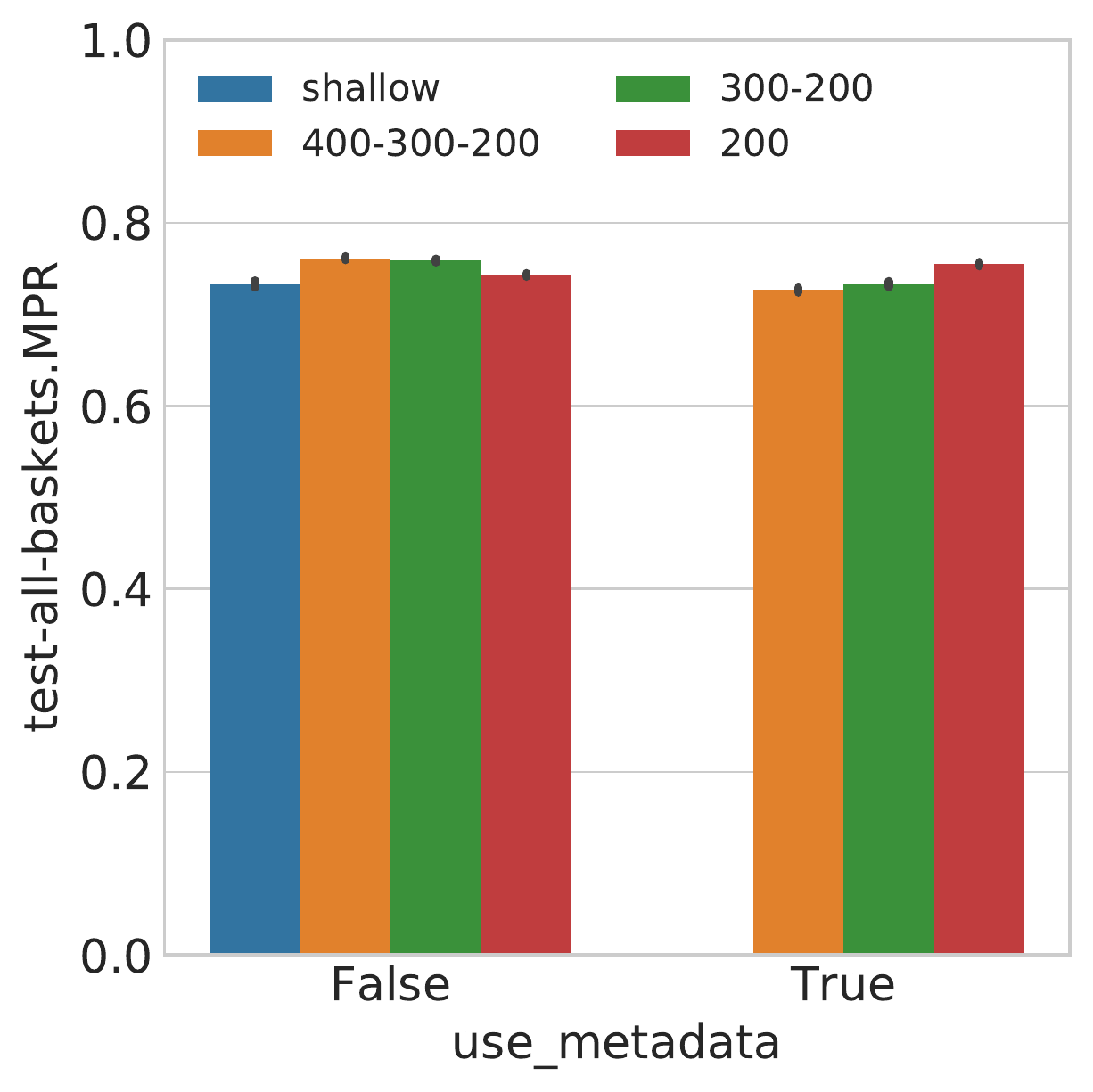}
    \end{subfigure}
    \begin{subfigure}{.24\textwidth}
        \centering
        \includegraphics[width=\textwidth]{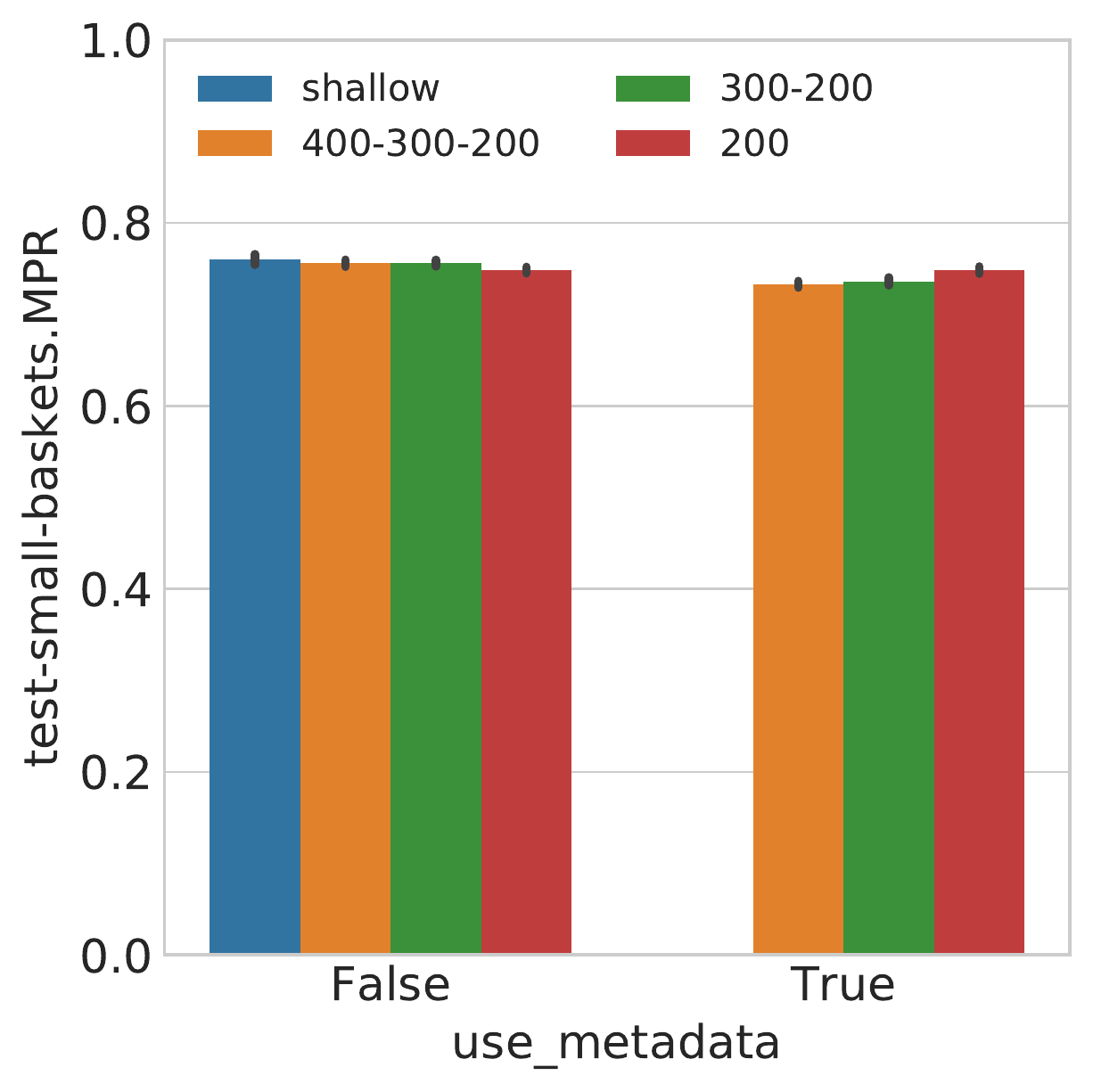}
    \end{subfigure}
    \begin{subfigure}{.24\textwidth}
        \centering
        \includegraphics[width=\textwidth]{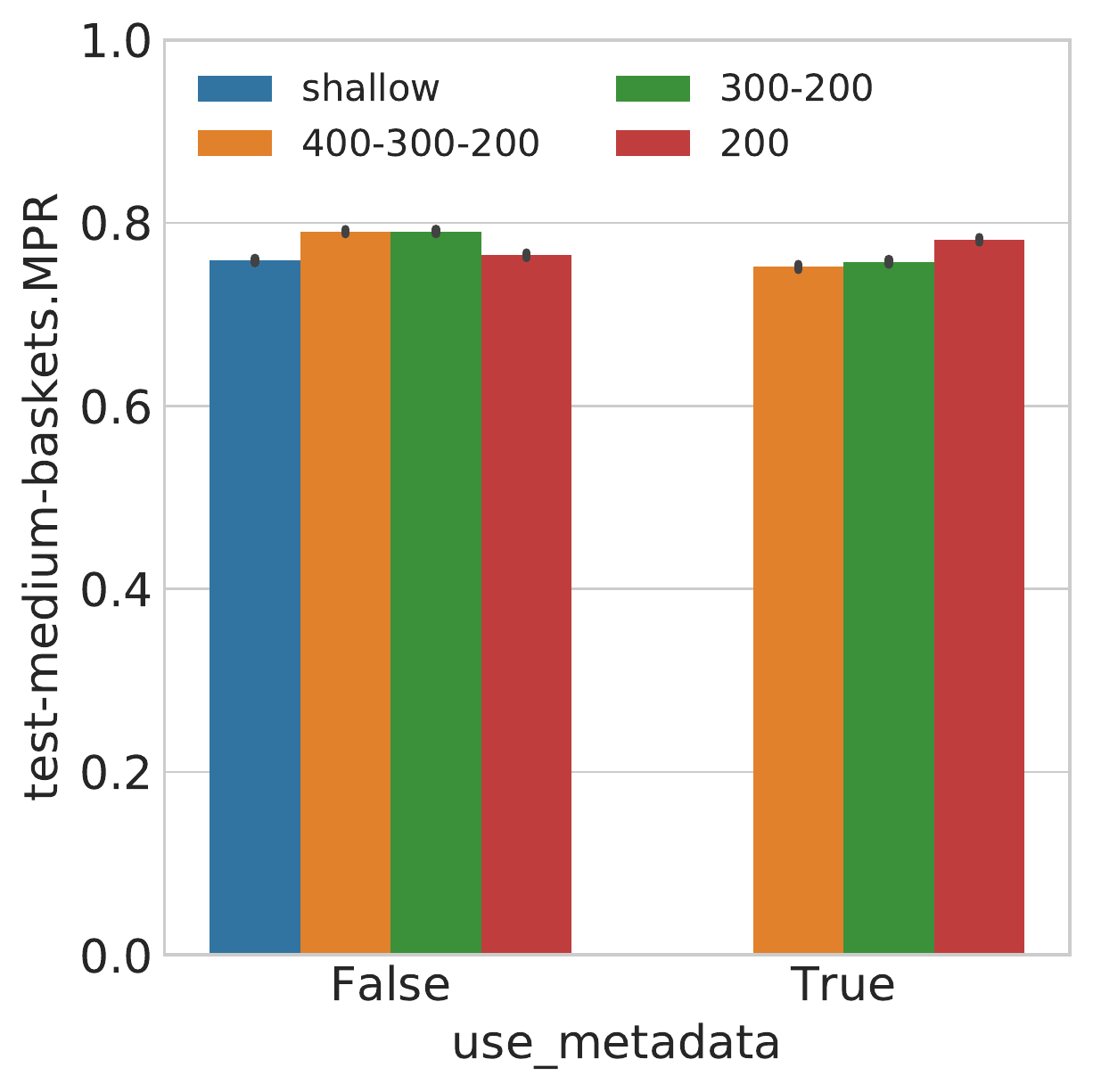}
    \end{subfigure}
    \begin{subfigure}{.24\textwidth}
        \centering
        \includegraphics[width=\textwidth]{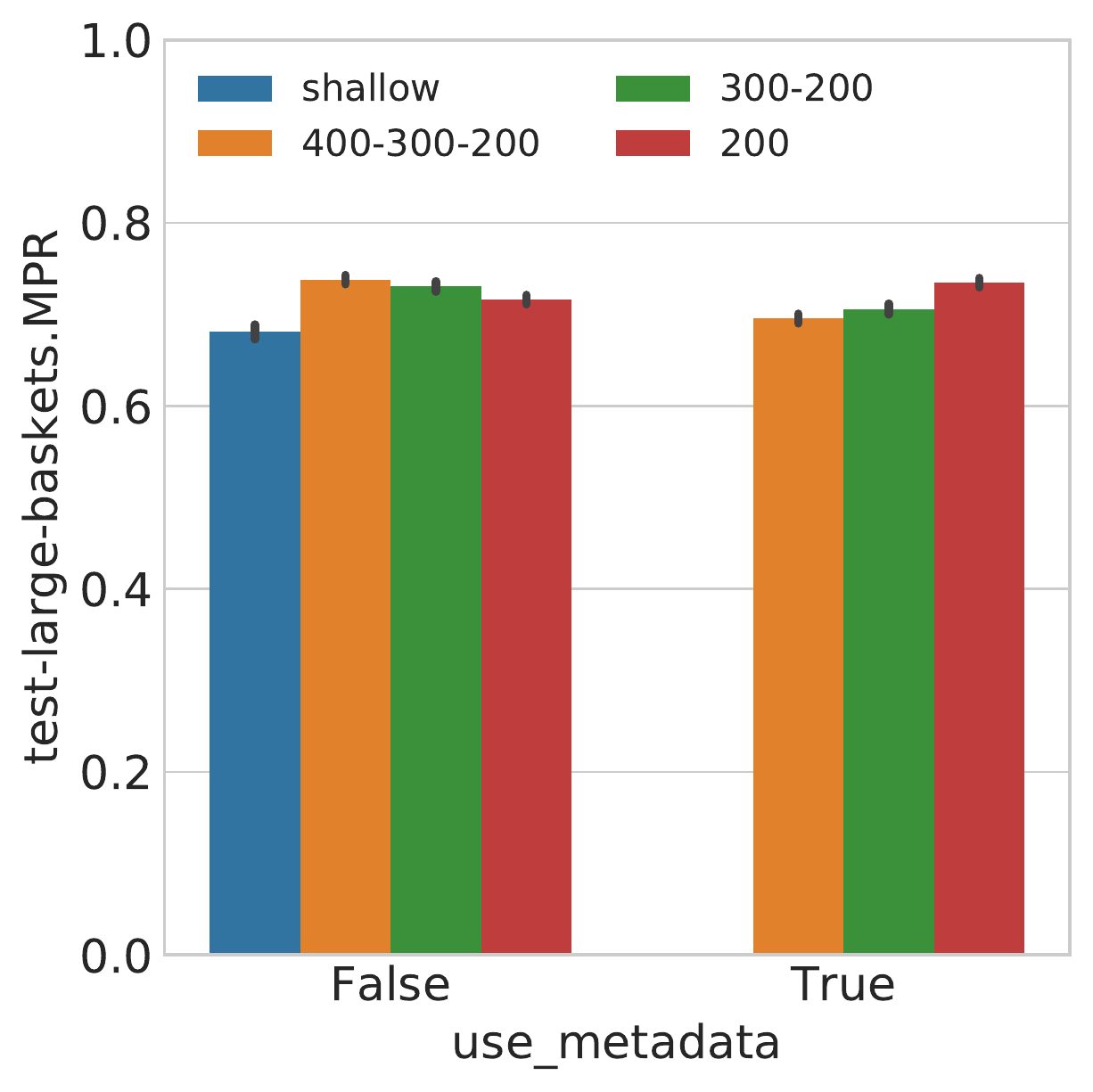}
    \end{subfigure}
  \vspace{-0.2cm}
  \caption{Instacart-10k MPR results, for shallow DPP and deep DPP models
  trained with and without metadata.  We show results for the shallow DPP model
  (the standard DPP, with no hidden layers), and for deep DPP models with one,
  two, and three hidden layers, denoted as 200, 300-200, and 400-300-200 hidden
  layer configurations, respectively.}
  \label{fig:instacart-10k-MPR-results}
  \vspace{-0.2cm}
\end{figure*}

\begin{figure*}[t]
    \centering
    \begin{subfigure}{.24\textwidth}
        \centering
        \includegraphics[width=\textwidth]{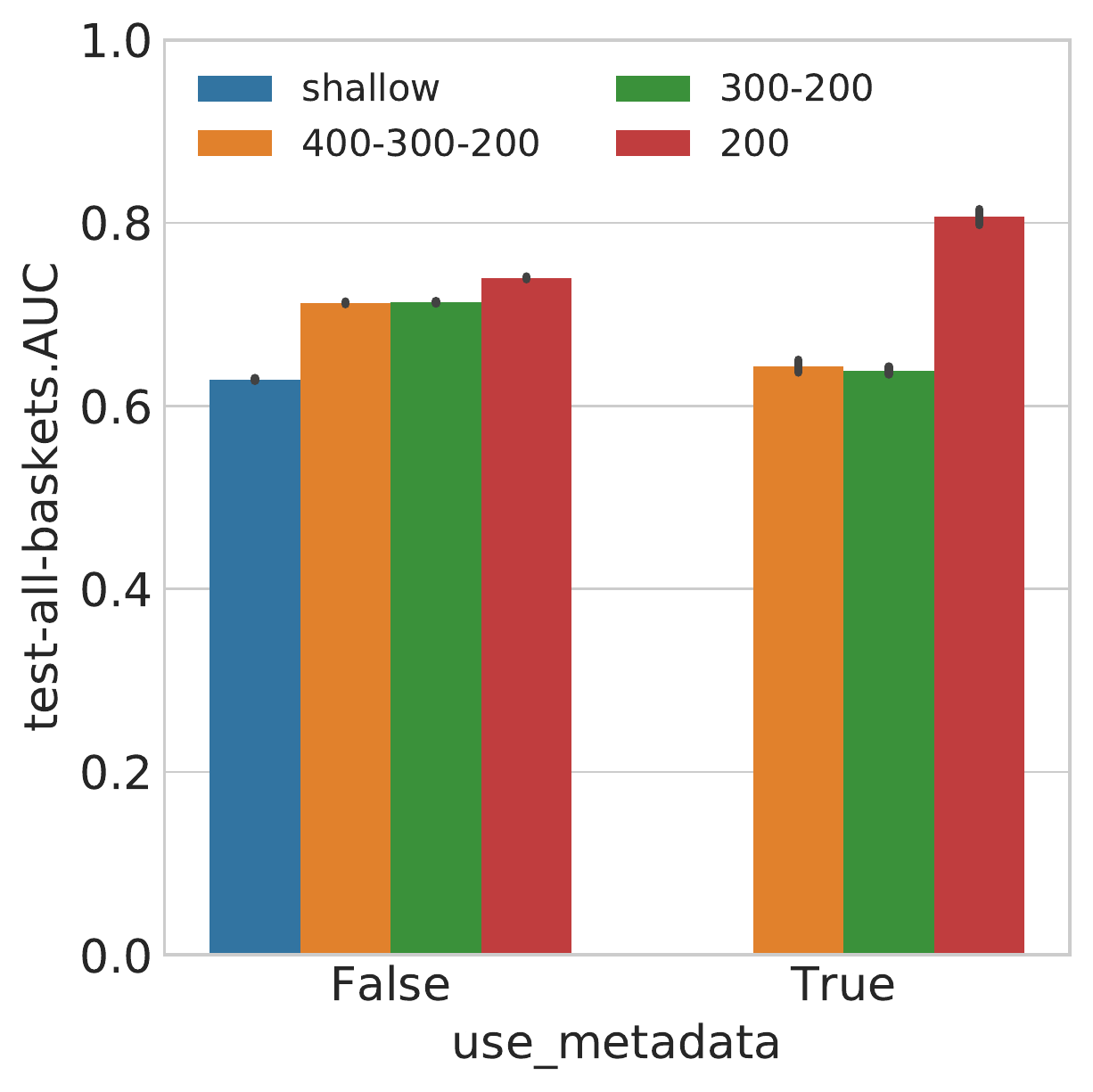}
    \end{subfigure}
    \begin{subfigure}{.24\textwidth}
        \centering
        \includegraphics[width=\textwidth]{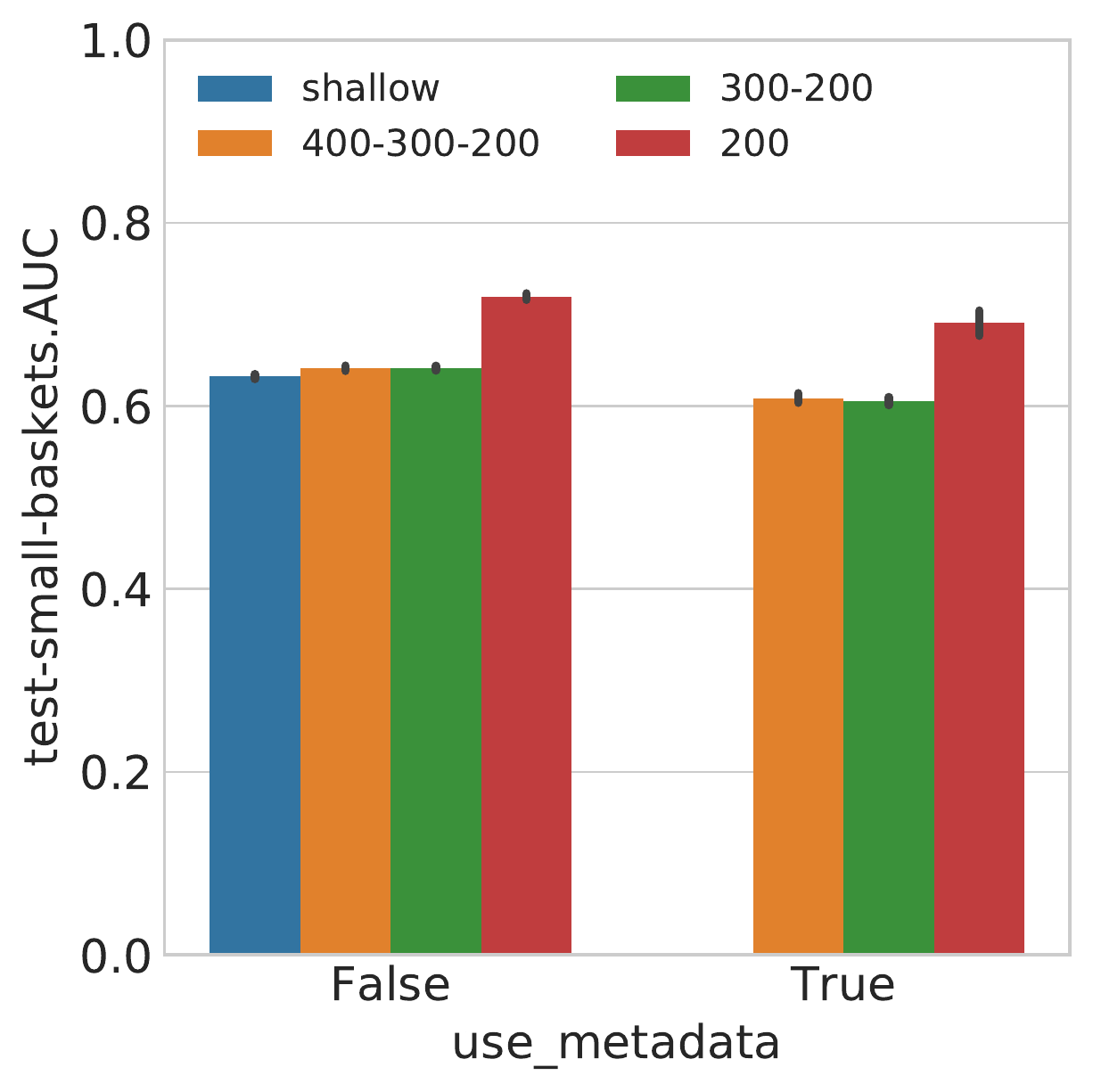}
    \end{subfigure}
    \begin{subfigure}{.24\textwidth}
        \centering
        \includegraphics[width=\textwidth]{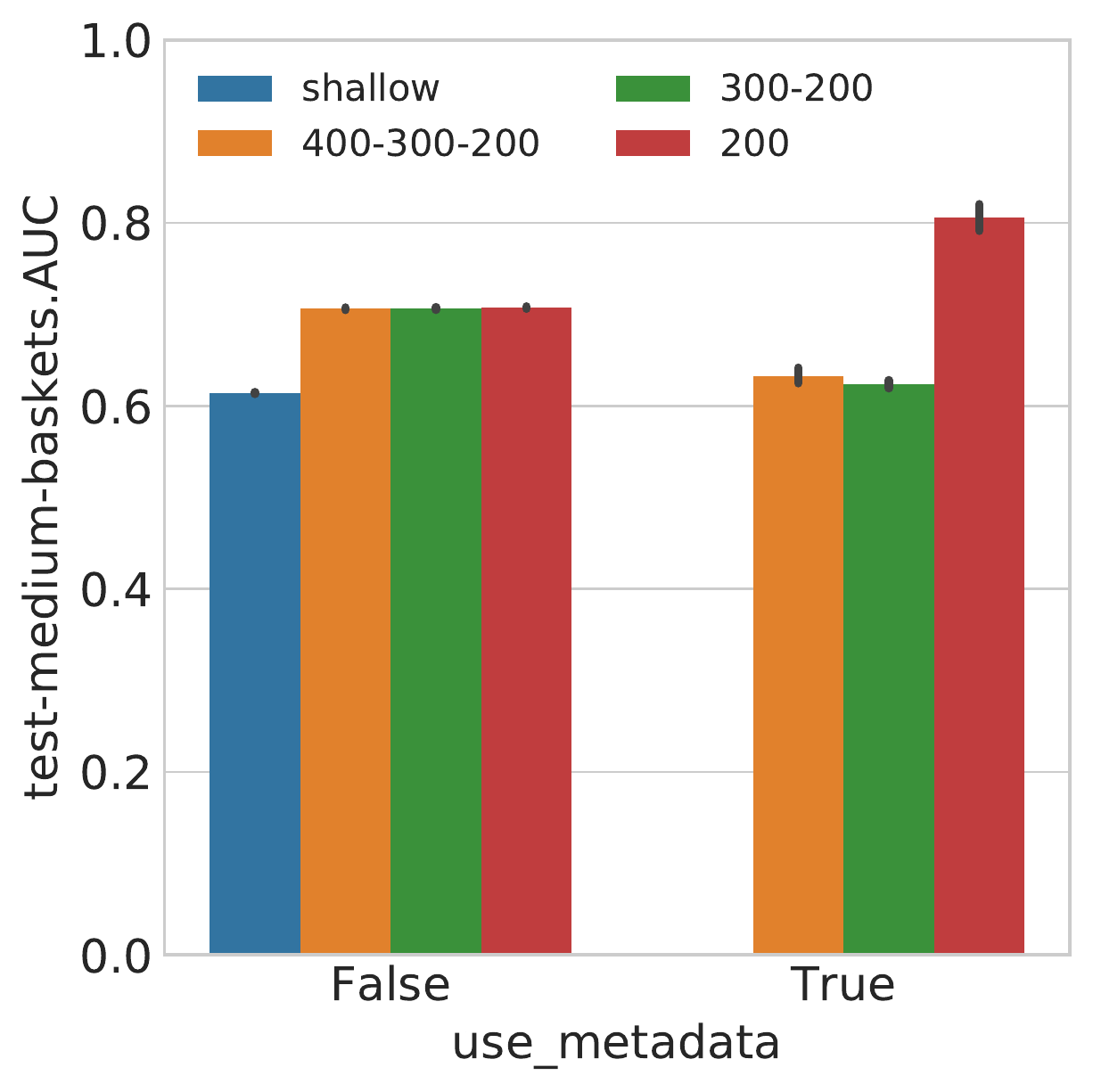}
    \end{subfigure}
    \begin{subfigure}{.24\textwidth}
        \centering
        \includegraphics[width=\textwidth]{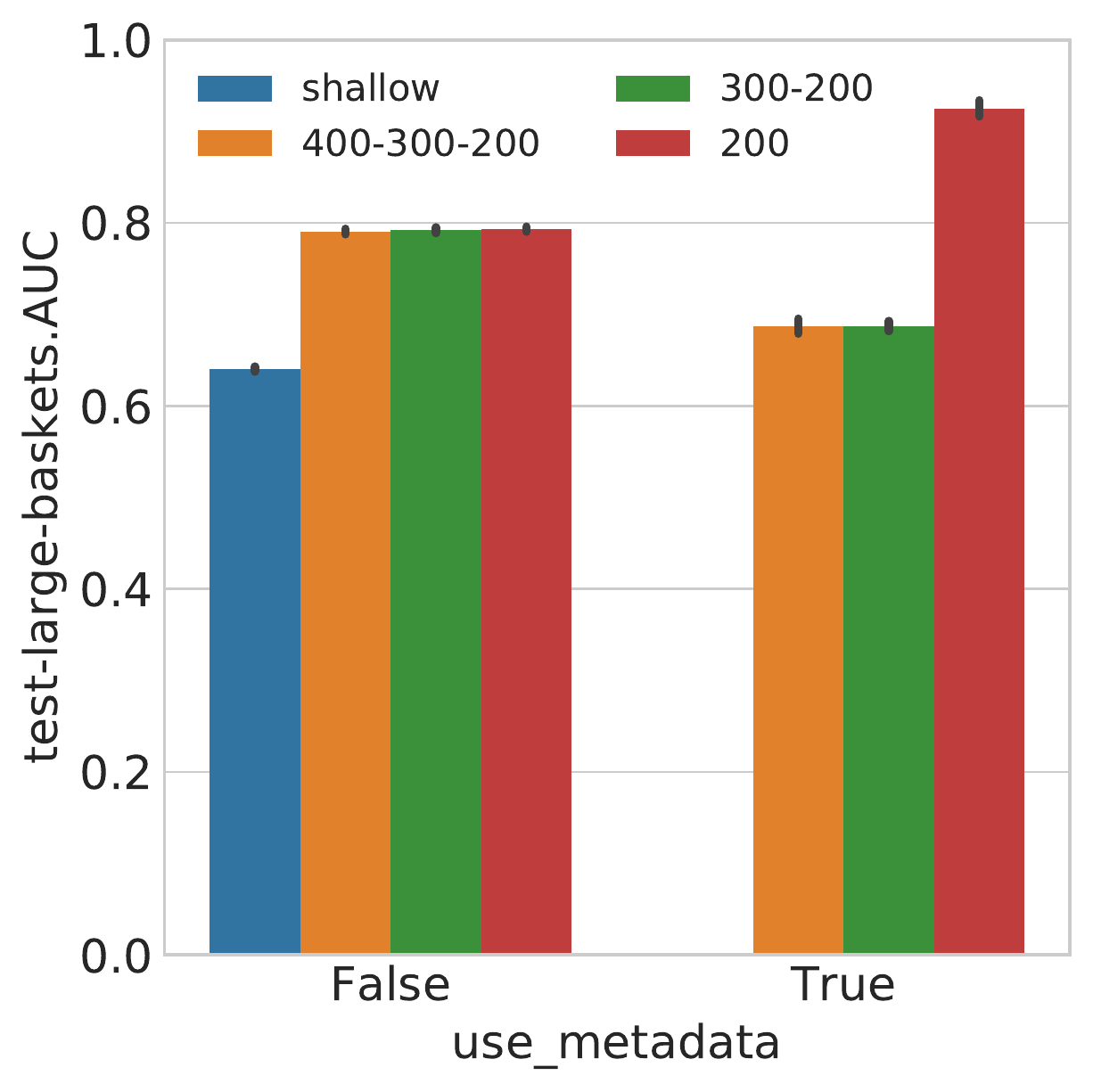}
    \end{subfigure}
  \vspace{-0.2cm}
  \caption{Instacart-10k AUC results, for shallow DPP and deep DPP models
  trained with and without metadata.  We show results for the shallow DPP model
  (the standard DPP, with no hidden layers), and for deep DPP models with one,
  two, and three hidden layers, denoted as 200, 300-200, and 400-300-200 hidden
  layer configurations, respectively.}
  \label{fig:instacart-10k-AUC-results}
  \vspace{-0.2cm}
\end{figure*}

\vspace{-0.2cm}
\subsection{Experimental setup and metrics}
\label{subsec:experimental-setup}

We compare the performance of all methods using a standard
recommender system metric: mean percentile rank (MPR).

We begin our definition of MPR by defining percentile rank (PR).  First,
given a set $A$, let $p_{i,A}=\Pr(A\cup\{i\} \mid A)$. The percentile rank of an
item $i$ given a set $A$ is defined as
\[\text{PR}_{i,A} = \frac {\sum_{i' \not\in A} \mathds 1(p_{i,A} \ge p_{i',A})}
{|\Ycal \wo A|} \times 100\%\]
where $\Ycal \wo A$ indicates those elements in
the ground set $\Ycal$ that are not found in $A$.

MPR is then computed as
\[\text{MPR}=\frac 1 {|\Tcal|} \sum_{A \in \mathcal
T}\text{PR}_{i,A\wo \{i\}}\] 
where $\Tcal$ is the set of test instances and $i$
is a randomly selected element in each set $A$. A MPR of 50 is equivalent to
random selection; a MPR of 100 indicates that the model perfectly predicts the
held out item.  MPR is a recall-based metric which we use to evaluate the
model's predictive power by measuring how well it predicts the next item in a
basket; it is a standard choice for recommender systems~\citep{hu08,li10}.





We evaluate the discriminative power of each model using the AUC metric. For
this task, we generate a set of negative subsets uniformly at random.  For each
positive subset $A^+$ in the test set, we generate a negative subset $A^-$ of
the same length by drawing $|A^+|$ samples uniformly at random, and ensure that
the same item is not drawn more than once for a subset.  We compute the AUC for
the model on these positive and negative subsets, where the score for each
subset is the log-likelihood that the model assigns to the subset. This task
measures the ability of the model to discriminate between observed positive
subsets (ground-truth subsets) and randomly generated subsets.

For all experiments, a random selection of 2000 baskets are used for testing, a
random selection of 300 baskets are used for the validation set for tuning
hyperparameters and tracking convergence, and the remaining baskets in the dataset
are used for training. Convergence is reached during training when the relative
change in validation log-likelihood is below a pre-determined threshold, which
is set identically for all models; or after training for a maximum of 1000
iterations.  We set $\alpha = 1$ for the standard low-rank DPP baseline model,
which we found to be a reasonably optimal value for all datasets used in our
evaluation, in line with prior work~\cite{gartrell17}. We set $\alpha = 0$ for
the deep DPP models, which we found to be reasonably optimal for all datasets
and model configurations.

\begin{figure*}[t!]
    \centering
    \begin{subfigure}{.32\textwidth}
        \centering
        \includegraphics[width=\textwidth]{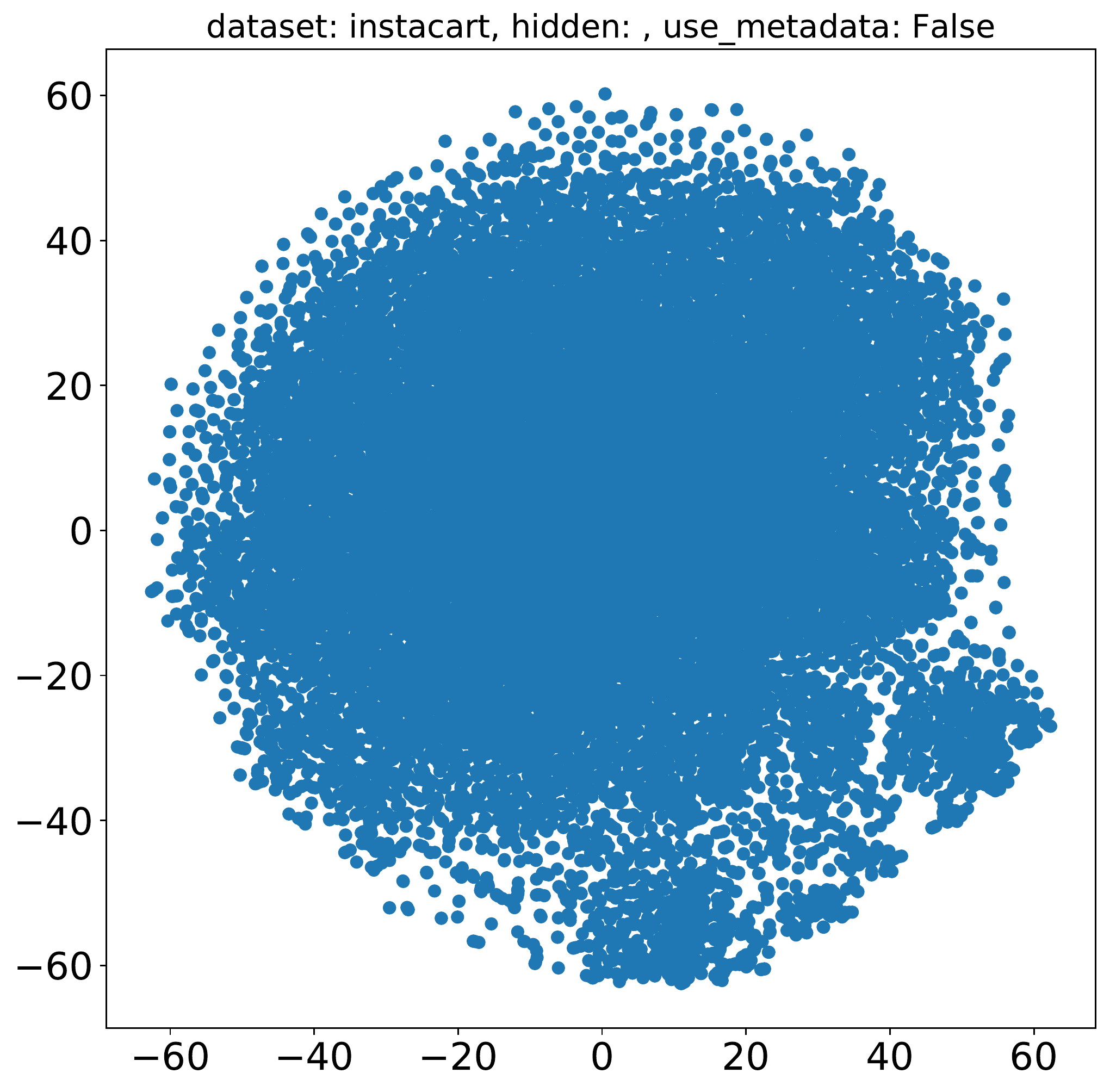}
        \caption{Shallow DPP}
    \end{subfigure}
    \begin{subfigure}{.32\textwidth}
        \centering
        \includegraphics[width=\textwidth]{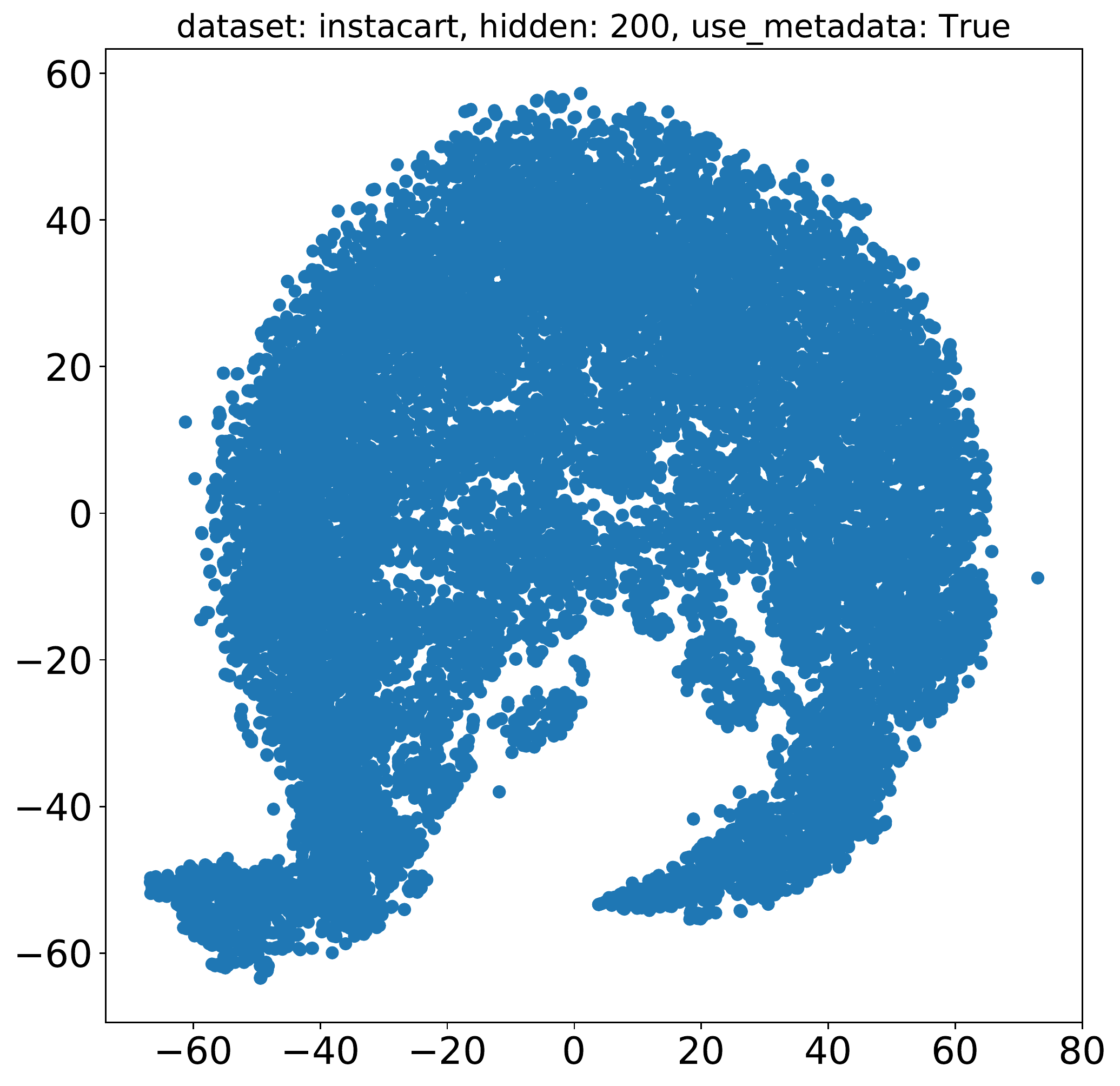}
        \caption{Deep DPP with 1 hidden layer}
    \end{subfigure}
    \begin{subfigure}{.31\textwidth}
        \centering
        \includegraphics[width=\textwidth]{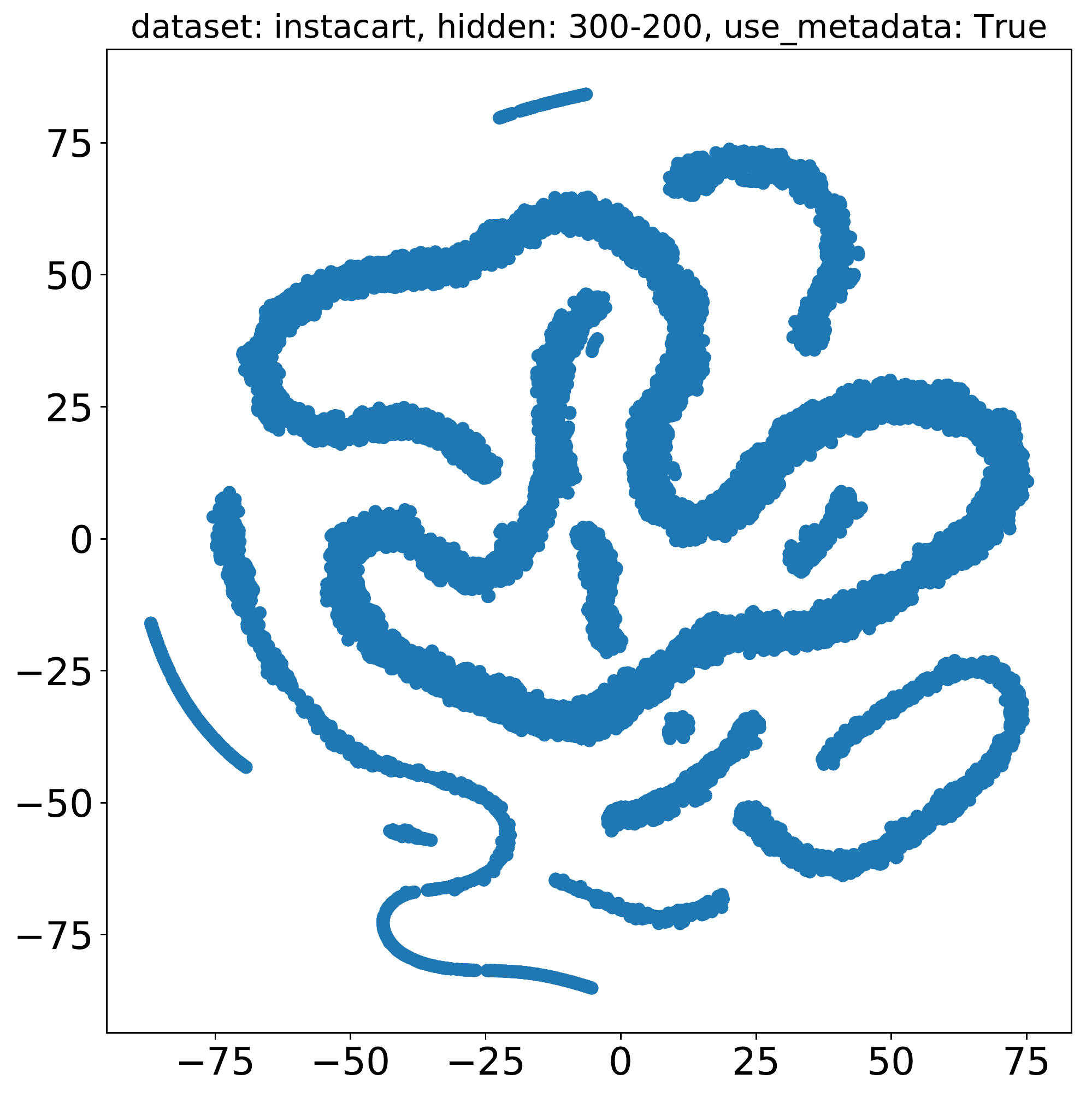}
        \caption{Deep DPP with 2 hidden layers}
    \end{subfigure}
  \vspace{-0.1cm}
  \caption{t-SNE plots of product embeddings for the Instacart-10k dataset, for
  the shallow DPP (with no hidden layers), deep DPP with one hidden layer, and deep
  DPP with two hidden layers.  The deep DPP models are trained with metadata, while
  the shallow DPP does not support metadata.}
  \label{fig:t-SNE}
  \vspace{-0.5cm}
\end{figure*}

\vspace{-0.2cm}
\subsection{Results}
\label{subsec:experimental-results}
\vspace{-0.2cm}
Figures~\ref{fig:MPR-results} through~\ref{fig:instacart-10k-AUC-results} show
the results of our experiments; these plots show mean and 95\% confidence
interval estimates (as error bars) for the MPR and AUC metrics obtained using
bootstrapping.  For Figures~\ref{fig:MPR-results} and~\ref{fig:AUC-results}, we
select optimal deep DPP model configuration across the results for deep DPPs
with one, two, and three hidden layers. The results for the shallow model in
these figures refers to the standard low-rank DPP model, with 0 hidden layers.
In addition to computing our results on all test baskets, we also computed
results on the test set divided into three equally-sized populations segmented
according to basket size.

Figure~\ref{fig:MPR-results} shows the MPR results for our experiments. Compared
to the best performing baseline models, we see that the deep DPP model leads to
larger MPR improvements for the Instacart and Belgian datasets than for the UK
dataset (where the PF model performs best). The Instacart and Belgian datasets
are of higher complexity than the UK datasets, so the larger improvement in MPR
for these two datasets suggests that the deep DPP is able to effectively capture
the additional signal available in higher complexity datasets.  The deep DPP
substantially outperforms the pre-trained DPP ("pre-trained Poisson") in most
cases, indicating the value of end-to-end learning of the deep DPP kernel as
compared to building the kernel from pre-trained embeddings obtained from
another model.

Figure~\ref{fig:AUC-results} shows the AUC results for our experiments.  We see
that the deep DPP model provides moderate to large AUC improvements for the
Instacart, Belgian, and UK datasets over the standard low-rank DPP.  Compared to
the low-rank DPP, the deep DPP is above to provide somewhat larger AUC
improvements for the Instacart and Belgian datasets than for the UK dataset,
again suggesting that the deep DPP is able capture the additional complexity in
interaction among items in higher complexity datasets.  Overall, these results
suggest that the using a deep network to learn the DPP kernel is effective at
improving the discriminative power of DPPs, particularly for more complex
datasets.  We see that the deep DPP matches or outperforms the pre-trained DPP
in most cases (expect for the Belgian dataset), providing further evidence of
the value of end-to-end learning of the deep DPP kernel.  Note that we cannot
use the PF model to assign a score to a subset (basket) in a straightforward
manner, and so we omit the PF model from this analysis.

To evaluate the impact of training deep DPP models with metadata,
figures~\ref{fig:instacart-10k-MPR-results}
and~\ref{fig:instacart-10k-AUC-results} show the MPR and AUC results for our
experiments on the Instacart-10k dataset, where models are trained with and
without the metadata available for this dataset.  We use the Instacart-10k
dataset for these experiments since it has higher sparsity than the full
Instacart dataset, emphasizing the impact of metadata in a high-sparsity
setting. The shallow (standard) low-rank DPP is unable to natively support
metadata, since this would require manual feature engineering that we have not
implemented, and hence results for this model with metadata are not available.
From the AUC results, we see that training deep DPP models with metadata is
effective at significantly improving the discriminative power of the model. From
the MPR results, we see that using metadata generally provides small to moderate
improvements in predictive performance.  We see somewhat larger MPR improvements
for large baskets when using metadata.  Since larger baskets are less common in
this dataset, this result suggests that metadata can be useful in improving
model performance for sparse regions of the data.  The substantial AUC
improvement of approximately 0.2 for large baskets when using metadata for the
best deep model, as compared to the shallow DPP without metadata, is another
indication of the improvements that metadata can bring for sparse areas of the
data.

t-SNE plots~\cite{maaten2008visualizing} for the item embeddings learned by the
deep DPP and standard DPP for the Instacart-10k dataset are shown in
Figure~\ref{fig:t-SNE}.  We see that the standard shallow DPP learns item
embeddings that are approximately distributed within a sphere.  For the deep DPP
models, we see that as we increase the number of hidden layers, the structure of
the embedding space becomes more well defined. These plots suggest that deep DPP
models are able to learn complex nonlinear interactions between items.

\vspace{-0.3cm}
\section{Conclusion}
\label{sec:conclusion}
\vspace{-0.3cm}
We have introduced the deep DPP model, which uses a deep feed-forward neural
network to learn the DPP kernel matrix containing item embeddings.  The deep DPP
overcomes several limitations of the standard DPP model by allowing us to
arbitrary increase the expressive power of the model through capturing nonlinear
item interactions, while still leveraging the efficient learning, sampling, and
prediction algorithms available for standard DPPs.  The deep DPP architecture
also allows us to easily incorporate item metadata into DPP learning.
Experimentally, we have shown that compared to standard DPPs, which can only
capture linear interactions among items, the deep DPP can significantly improve
predictive performance, while also improving the model's ability to discriminate
between real and randomly generated subsets.  Our evaluation also shows that the
deep DPP is capable of outperforming strong baseline models in many cases.  




\bibliographystyle{plainnat}
\setcitestyle{numbers}
\bibliography{bibliography}

\begin{thebibliography}{42}
\providecommand{\natexlab}[1]{#1}
\providecommand{\url}[1]{\texttt{#1}}
\expandafter\ifx\csname urlstyle\endcsname\relax
  \providecommand{\doi}[1]{doi: #1}\else
  \providecommand{\doi}{doi: \begingroup \urlstyle{rm}\Url}\fi

\bibitem[Affandi et~al.(2014)Affandi, Fox, Adams, and Taskar]{affandi14}
R.~Affandi, E.~Fox, R.~Adams, and B.~Taskar.
\newblock Learning the parameters of determinantal point process kernels.
\newblock In \emph{ICML}, 2014.

\bibitem[Bojanowski et~al.(2017)Bojanowski, Grave, Joulin, and
  Mikolov]{bojanowski2017enriching}
Piotr Bojanowski, Edouard Grave, Armand Joulin, and Tomas Mikolov.
\newblock Enriching word vectors with subword information.
\newblock \emph{Transactions of the Association for Computational Linguistics},
  5:\penalty0 135--146, 2017.
\newblock ISSN 2307-387X.

\bibitem[Borodin(2009)]{borodin2009}
Alexei Borodin.
\newblock {D}eterminantal {P}oint {P}rocesses.
\newblock \emph{arXiv:0911.1153}, 2009.

\bibitem[Brijs(2003)]{brijs03}
Tom Brijs.
\newblock Retail market basket data set.
\newblock In \emph{Workshop on Frequent Itemset Mining Implementations
  (FIMI’03)}, 2003.

\bibitem[Brijs et~al.(1999)Brijs, Swinnen, Vanhoof, and Wets]{brijs99}
Tom Brijs, Gilbert Swinnen, Koen Vanhoof, and Geert Wets.
\newblock Using association rules for product assortment decisions: A case
  study.
\newblock In \emph{Proceedings of the fifth ACM SIGKDD international conference
  on Knowledge discovery and data mining}, pages 254--260. ACM, 1999.

\bibitem[Chao et~al.(2015)Chao, Gong, Grauman, and Sha]{chao15}
Wei{-}Lun Chao, Boqing Gong, Kristen Grauman, and Fei Sha.
\newblock Large-margin determinantal point processes.
\newblock In \emph{Uncertainty in Artificial Intelligence (UAI)}, 2015.

\bibitem[Chen(2012)]{lsbupr1492}
D~Chen.
\newblock Data mining for the online retail industry: A case study of rfm
  model-based customer segmentation using data mining.
\newblock \emph{Journal of Database Marketing and Customer Strategy
  Management}, 19\penalty0 (3), August 2012.

\bibitem[Covington et~al.(2016)Covington, Adams, and Sargin]{covington2016deep}
Paul Covington, Jay Adams, and Emre Sargin.
\newblock Deep neural networks for youtube recommendations.
\newblock In \emph{RecSys}. ACM, 2016.

\bibitem[Decreusefond et~al.(2015)Decreusefond, Flint, Privault, and
  Torrisi]{decreuse}
Laurent Decreusefond, Ian Flint, Nicolas Privault, and Giovanni~Luca Torrisi.
\newblock {D}eterminantal {P}oint {P}rocesses, 2015.

\bibitem[Dupuy and Bach(2016)]{dupuy16}
Christophe Dupuy and Francis Bach.
\newblock Learning {D}eterminantal {P}oint {P}rocesses in sublinear time, 2016.

\bibitem[Gartrell et~al.(2016)Gartrell, Paquet, and
  Koenigstein]{gartrell2016bayesian}
Mike Gartrell, Ulrich Paquet, and Noam Koenigstein.
\newblock Bayesian low-rank determinantal point processes.
\newblock In \emph{RecSys}. ACM, 2016.

\bibitem[Gartrell et~al.(2017)Gartrell, Paquet, and Koenigstein]{gartrell17}
Mike Gartrell, Ulrich Paquet, and Noam Koenigstein.
\newblock Low-rank factorization of {D}eterminantal {P}oint {P}rocesses.
\newblock In \emph{AAAI}, 2017.

\bibitem[Gillenwater(2014)]{gillenwater-thesis}
J.~Gillenwater.
\newblock \emph{{Approximate Inference for {D}eterminantal {P}oint
  {P}rocesses}}.
\newblock PhD thesis, University of Pennsylvania, 2014.

\bibitem[Gillenwater et~al.(2014)Gillenwater, Kulesza, Fox, and
  Taskar]{gillenwater14}
J.~Gillenwater, A.~Kulesza, E.~Fox, and B.~Taskar.
\newblock Expectation-maximization for learning {D}eterminantal {P}oint
  {P}rocesses.
\newblock In \emph{NIPS}, 2014.

\bibitem[Gopalan et~al.(2015)Gopalan, Hofman, and Blei]{gopalan2015}
Prem Gopalan, Jake~M Hofman, and David~M Blei.
\newblock Scalable recommendation with hierarchical {P}oisson factorization.
\newblock In \emph{UAI}, 2015.

\bibitem[Hu et~al.(2008)Hu, Koren, and Volinsky]{hu08}
Yifan Hu, Yehuda Koren, and Chris Volinsky.
\newblock Collaborative filtering for implicit feedback datasets.
\newblock In \emph{Proceedings of the 2008 Eighth IEEE International Conference
  on Data Mining}, 2008.

\bibitem[Kingma and Ba(2015)]{kingma2015adam}
Diederik~P Kingma and Jimmy Ba.
\newblock Adam: A method for stochastic optimization.
\newblock In \emph{ICLR}, 2015.

\bibitem[Klambauer et~al.(2017)Klambauer, Unterthiner, Mayr, and
  Hochreiter]{klambauer2017self}
G{\"u}nter Klambauer, Thomas Unterthiner, Andreas Mayr, and Sepp Hochreiter.
\newblock Self-normalizing neural networks.
\newblock In \emph{NIPS}, 2017.

\bibitem[Krause et~al.(2008)Krause, Singh, and Guestrin]{krause08}
Andreas Krause, Ajit Singh, and Carlos Guestrin.
\newblock Near-optimal sensor placements in {G}aussian processes: theory,
  efficient algorithms and empirical studies.
\newblock \emph{JMLR}, 9:\penalty0 235--284, 2008.

\bibitem[Kula(2015)]{kula2015metadata}
Maciej Kula.
\newblock Metadata embeddings for user and item cold-start recommendations.
\newblock \emph{arXiv preprint arXiv:1507.08439}, 2015.

\bibitem[Kulesza(2013)]{kuleszaThesis}
A.~Kulesza.
\newblock \emph{Learning with {D}eterminantal {P}oint {P}rocesses}.
\newblock PhD thesis, University of Pennsylvania, 2013.

\bibitem[Kulesza and Taskar(2011{\natexlab{a}})]{kulesza11}
A.~Kulesza and B.~Taskar.
\newblock k-dpps: Fixed-size determinantal point processes.
\newblock In \emph{ICML}, 2011{\natexlab{a}}.

\bibitem[Kulesza and Taskar(2012)]{kuleszaBook}
A.~Kulesza and B.~Taskar.
\newblock \emph{{D}eterminantal {P}oint {P}rocesses for machine learning},
  volume~5.
\newblock Foundations and Trends in Machine Learning, 2012.

\bibitem[Kulesza and Taskar(2011{\natexlab{b}})]{kulesza2011learning}
Alex Kulesza and Ben Taskar.
\newblock Learning determinantal point processes.
\newblock In \emph{UAI}, 2011{\natexlab{b}}.

\bibitem[Lavancier et~al.(2015)Lavancier, M{\o}ller, and Rubak]{lavancier15}
Fr{\'e}d{\'e}ric Lavancier, Jesper M{\o}ller, and Ege Rubak.
\newblock {D}eterminantal {P}oint {P}rocess models and statistical inference.
\newblock \emph{Journal of the Royal Statistical Society: Series B (Statistical
  Methodology)}, 77\penalty0 (4):\penalty0 853--877, 2015.

\bibitem[Li et~al.(2010)Li, Hu, Zhai, and Chen]{li10}
Yanen Li, Jia Hu, ChengXiang Zhai, and Ye~Chen.
\newblock Improving one-class collaborative filtering by incorporating rich
  user information.
\newblock In \emph{Proceedings of the 19th ACM International Conference on
  Information and Knowledge Management}, CIKM '10, 2010.

\bibitem[Lin and Bilmes(2012)]{lin12}
H.~Lin and J.~Bilmes.
\newblock Learning mixtures of submodular shells with application to document
  summarization.
\newblock In \emph{Uncertainty in Artificial Intelligence (UAI)}, 2012.

\bibitem[Maaten and Hinton(2008)]{maaten2008visualizing}
Laurens van~der Maaten and Geoffrey Hinton.
\newblock Visualizing data using t-sne.
\newblock \emph{Journal of machine learning research}, 9\penalty0
  (Nov):\penalty0 2579--2605, 2008.

\bibitem[Mariet and Sra(2015)]{mariet15}
Zelda Mariet and Suvrit Sra.
\newblock Fixed-point algorithms for learning {D}eterminantal {P}oint
  {P}rocesses.
\newblock In \emph{ICML}, 2015.

\bibitem[Mariet and Sra(2016{\natexlab{a}})]{mariet16}
Zelda Mariet and Suvrit Sra.
\newblock Diversity networks.
\newblock \emph{Int. Conf. on Learning Representations (ICLR)},
  2016{\natexlab{a}}.

\bibitem[Mariet and Sra(2016{\natexlab{b}})]{mariet16b}
Zelda Mariet and Suvrit Sra.
\newblock Kronecker {D}eterminantal {P}oint {P}rocesses.
\newblock In \emph{NIPS}, 2016{\natexlab{b}}.

\bibitem[Mariet et~al.(2019{\natexlab{a}})Mariet, Gartrell, and
  Sra]{mariet2019negdpp}
Zelda Mariet, Mike Gartrell, and Suvrit Sra.
\newblock Learning determinantal point processes by sampling inferred
  negatives.
\newblock In \emph{AISTATS, to appear}, 2019{\natexlab{a}}.

\bibitem[Mariet et~al.(2019{\natexlab{b}})Mariet, Ovadia, and
  Snoek]{mariet2019dppnet}
Zelda Mariet, Yaniv Ovadia, and Jasper Snoek.
\newblock Dppnet: Approximating determinantal point processes with deep
  networks.
\newblock \emph{arXiv preprint arXiv:1901.02051}, 2019{\natexlab{b}}.

\bibitem[Osogami et~al.(2018)Osogami, Raymond, Goel, Shirai, and
  Maehara]{osogami18}
Takayuki Osogami, Rudy Raymond, Akshay Goel, Tomoyuki Shirai, and Takanori
  Maehara.
\newblock Dynamic {D}eterminantal {P}oint {P}rocesses.
\newblock In \emph{AAAI}, 2018.

\bibitem[Recht et~al.(2011)Recht, Re, Wright, and Niu]{recht2011hogwild}
Benjamin Recht, Christopher Re, Stephen Wright, and Feng Niu.
\newblock Hogwild: A lock-free approach to parallelizing stochastic gradient
  descent.
\newblock In \emph{NIPS}, 2011.

\bibitem[Urschel et~al.(2017)Urschel, Brunel, Moitra, and Rigollet]{urschel17}
John Urschel, Victor{-}Emmanuel Brunel, Ankur Moitra, and Philippe Rigollet.
\newblock Learning {D}eterminantal {P}oint {P}rocesses with moments and cycles.
\newblock In \emph{Proceedings of the 34th International Conference on Machine
  Learning, {ICML}}, pages 3511--3520, 2017.

\bibitem[Vasile et~al.(2016)Vasile, Smirnova, and Conneau]{vasile2016meta}
Flavian Vasile, Elena Smirnova, and Alexis Conneau.
\newblock Meta-prod2vec: Product embeddings using side-information for
  recommendation.
\newblock In \emph{RecSys}, 2016.

\bibitem[Wilhelm et~al.(2018)Wilhelm, Ramanathan, Bonomo, Jain, Chi, and
  Gillenwater]{wilhelm2018practical}
Mark Wilhelm, Ajith Ramanathan, Alexander Bonomo, Sagar Jain, Ed~H Chi, and
  Jennifer Gillenwater.
\newblock Practical diversified recommendations on youtube with determinantal
  point processes.
\newblock In \emph{CIKM}. ACM, 2018.

\bibitem[Xie et~al.(2017)Xie, Salakhutdinov, Mou, and Xing]{xie2017deep}
Pengtao Xie, Ruslan Salakhutdinov, Luntian Mou, and Eric~P Xing.
\newblock Deep determinantal point process for large-scale multi-label
  classification.
\newblock In \emph{ICCV}, pages 473--482, 2017.

\bibitem[Xue et~al.(2017)Xue, Dai, Zhang, Huang, and Chen]{xue2017deep}
Hong-Jian Xue, Xinyu Dai, Jianbing Zhang, Shujian Huang, and Jiajun Chen.
\newblock Deep matrix factorization models for recommender systems.
\newblock In \emph{IJCAI}, 2017.

\bibitem[Zaheer et~al.(2017)Zaheer, Kottur, Ravanbakhsh, Poczos, Salakhutdinov,
  and Smola]{zaheer2017deep}
Manzil Zaheer, Satwik Kottur, Siamak Ravanbakhsh, Barnabas Poczos, Ruslan~R
  Salakhutdinov, and Alexander~J Smola.
\newblock Deep sets.
\newblock In \emph{NIPS}, pages 3391--3401, 2017.

\bibitem[Zhang et~al.(2017)Zhang, Kjellstr{\"{o}}m, and Mandt]{zhang17}
Cheng Zhang, Hedvig Kjellstr{\"{o}}m, and Stephan Mandt.
\newblock Stochastic learning on imbalanced data: {D}eterminantal {P}oint
  {P}rocesses for mini-batch diversification.
\newblock \emph{CoRR}, abs/1705.00607, 2017.

\end{thebibliography}

\clearpage
\appendix

\section{Computing Predictions}
\label{section:computing-predictions}
Next-item prediction involves identifying the best item to add to a subset of
chosen objects (e.g., basket completion), and is the primary prediction task we
evaluate in Section~\ref{sec:experiments}. We compute next-item predictions for
subsets using the approach for efficient low-rank DPP conditioning described
in~\cite{mariet2019negdpp}. As in~\cite{gillenwater-thesis}, we first compute
the dual kernel $\C = \B^\top\B$, where $\B = \V^\top$.  We then compute
\begin{equation*}
 \C^A = (\B^A)^\top \B^A = \Z^A \C \Z^A,
\end{equation*}
with $\Z^A = \I - \B_A(\B_A^\top \B_A)^{-1} \B_A^\top$, where $\C^A$ is the
DPP kernel conditioned on the event that all items in $A$ are observed, and
$\B_A$ is the restriction of $\B$ to the rows and columns indexed by $A$.

Next, as described in~\cite{kuleszaBook}, we eigendecompose $\C^A$ to compute
the conditional (marginal) probability $P_i$ of every possible item $i$ in
$\bar{A}$:
\begin{equation*}
P_i = \sum\nolimits_{n=1}^K \tfrac{\lambda_n}{\lambda_n + 1}\left( \tfrac{1}{\sqrt{\lambda_n}} \b_i^A \hat{\v}_n  \right)^2
\end{equation*}
where $\b_i^A$ is a column vector for item $i$ in $\B^A$, $(\lambda_n$, $\hat
\v_n)$ are an eigenvalue/vector of $\C^A$, and $\bar{A} = \Ycal - A$.  As shown
in~\cite{mariet2019negdpp}, the overall computational complexity for computing
next-item conditionals/predictions for the low-rank DPP using this dual kernel
approach is $\mathcal O(K^3 + |A|^3 + K^2 |A|^2 + |\bar{A}| K^2)$.  Since in
most cases $K \ll |\bar{A}|$, this allows for efficient conditioning that is
essentially linear in the size of the catalog.

\end{document}